%% file: main.tex
\documentclass[10pt,twocolumn,letterpaper]{article}

\usepackage[pagenumbers]{cvpr} 
\usepackage{graphicx}
\usepackage{amsmath}
\usepackage{amssymb}
\usepackage{array}
\usepackage{booktabs}
\usepackage{multirow}
\usepackage{threeparttable}
\usepackage{bbding}
\usepackage{makecell}
\usepackage{subcaption}
\usepackage[table]{xcolor}
\usepackage{bm}
\usepackage{bbm}
\usepackage[pagebackref,breaklinks,colorlinks]{hyperref}
\newlength\savewidth\newcommand\shline{\noalign{\global\savewidth\arrayrulewidth
  \global\arrayrulewidth 1pt}\hline\noalign{\global\arrayrulewidth\savewidth}}
\newcommand\paperurl[1]{{\footnotesize{\color{blue}{\url{#1}}}}}
\usepackage{booktabs}
\usepackage{pifont}
\newcommand{\cmark}{\ding{51}}%
\newcommand{\xmark}{\ding{55}}%
\usepackage[capitalize]{cleveref}
\crefname{section}{Sec.}{Secs.}
\Crefname{section}{Section}{Sections}
\Crefname{table}{Table}{Tables}
\crefname{table}{Tab.}{Tabs.}

\input{preamble}
\begin{document}
\input{0_metadata}
\maketitle
\input{0_abstract}
\input{1_introduction}

% \clearpage
\input{2_related}

\input{3_method}
\input{4_results}
\input{5_conclusions}

%%%%%%%%% REFERENCES
{
    % \clearpage
    \small
    \bibliographystyle{ieee_fullname}
    \bibliography{macros,main}
}
% --- supplementary material
\clearpage
\input{X_supplementary}

% --- uncomment this to read the instructions
% \input{X_instructions}

\end{document}

%% file: preamble.tex
\usepackage{overpic}
\usepackage{enumitem} %< control spacing in itemize/enumerate/...
\usepackage{overpic} %< add raw math symbols to figures
\usepackage{color}
\definecolor{turquoise}{cmyk}{0.65,0,0.1,0.3}
\definecolor{purple}{rgb}{0.65,0,0.65}
\definecolor{dark_green}{rgb}{0, 0.5, 0}
\definecolor{orange}{rgb}{0.8, 0.6, 0.2}
\definecolor{red}{rgb}{0.8, 0.2, 0.2}
\definecolor{darkred}{rgb}{0.6, 0.1, 0.05}
\definecolor{blueish}{rgb}{0.0, 0.3, .6}
\definecolor{light_gray}{rgb}{0.7, 0.7, .7}
\definecolor{pink}{rgb}{1, 0, 1}
\definecolor{greyblue}{rgb}{0.25, 0.25, 1}

\usepackage{blindtext}

\renewcommand{\paragraph}[1]{\vspace{1em}\noindent\textbf{#1}.}

%% file: 0_metadata.tex
\title{DETRs with Collaborative Hybrid Assignments Training}

\author{
Zhuofan Zong
\quad Guanglu Song
\quad Yu Liu\thanks{Corresponding author.}
\\
{SenseTime Research} \\
\small{\texttt{\{zongzhuofan,liuyuisanai\}@gmail.com}} \\
\small{\texttt{songguanglu@sensetime.com}} 
}

%% file: 0_abstract.tex
\begin{abstract}
In this paper, 
we provide the observation that too few queries assigned as positive samples in DETR with one-to-one set matching leads to sparse supervision on the encoder's output which
considerably hurt the discriminative feature learning of the encoder and vice visa for attention learning in the decoder.
To alleviate this, we present a novel collaborative hybrid assignments training scheme, namely $\mathcal{C}$o-DETR, to learn more efficient and effective DETR-based detectors from versatile label assignment manners.
This new training scheme can easily enhance the encoder's learning ability in end-to-end detectors by training the multiple parallel auxiliary heads supervised by one-to-many label assignments such as ATSS and Faster RCNN.
In addition, we conduct extra customized positive queries by extracting the positive coordinates from these auxiliary heads to improve the training efficiency of positive samples in the decoder.
In inference, these auxiliary heads are discarded and thus our method introduces no additional parameters and computational cost to the original detector while requiring no hand-crafted non-maximum suppression (NMS).
We conduct extensive experiments to evaluate the effectiveness of the proposed approach on DETR variants, including DAB-DETR, Deformable-DETR, and DINO-Deformable-DETR.
% Specifically, we improve the basic Deformable-DETR by 5.8\% AP in 12-epoch training and 3.2\% AP in 36-epoch training. 
The state-of-the-art DINO-Deformable-DETR with Swin-L can be improved from 58.5\% to 59.5\% AP on COCO val.
Surprisingly, incorporated with ViT-L backbone, we achieve 66.0\% AP on COCO test-dev and 67.9\% AP on LVIS val, outperforming previous methods by clear margins with much fewer model sizes.
Codes are available at \url{https://github.com/Sense-X/Co-DETR}.
\end{abstract}

%% file: 1_introduction.tex
\section{Introduction}
\label{sec:intro}
\begin{figure}[t] 
    \centering
    \includegraphics[width=0.8\linewidth]{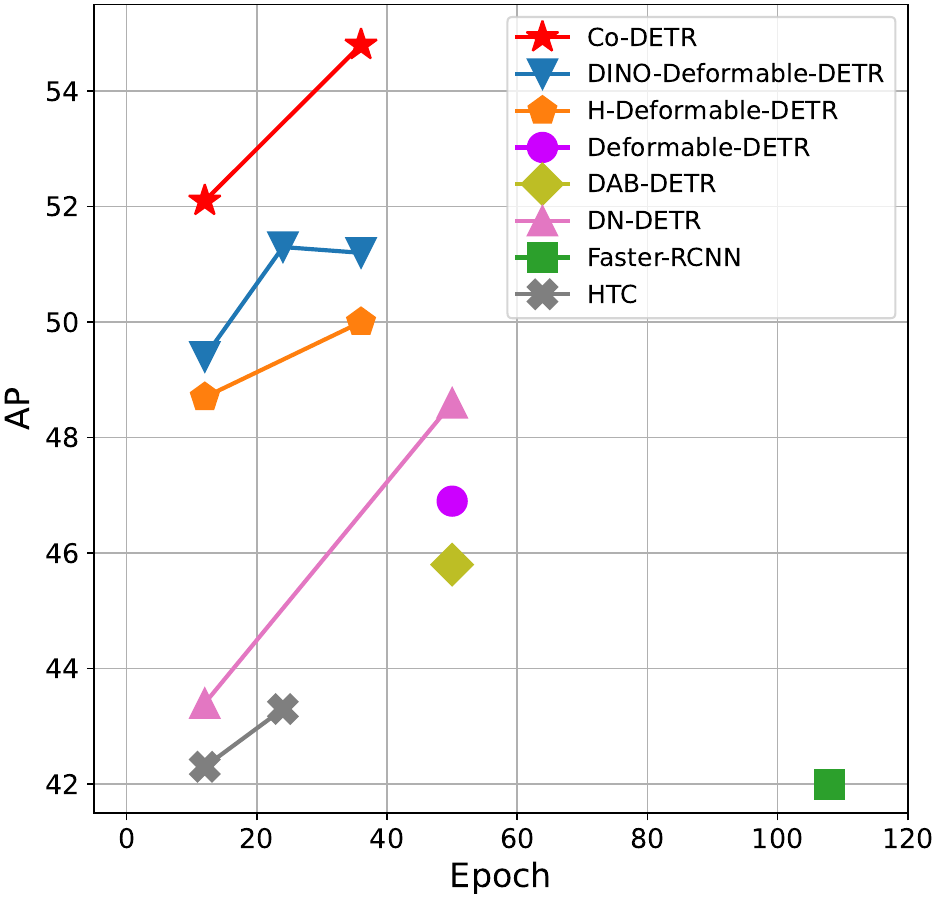}
    \vspace{-0.2cm}
    \caption{Performance of models with ResNet-50 on COCO \texttt{val}. $\mathcal{C}$o-DETR outperforms other counterparts by a large margin.
    % Note that DINO uses $900$ queries while $\mathcal{C}$o-Deformable-DETR uses $300$ queries during evaluation.
    }
    \label{fig:performance}
    \vspace{-0.3cm}
\end{figure}
Object detection is a fundamental task in computer vision, which requires us to localize the object and classify its category.
The seminal R-CNN families~\cite{fast,faster,mask} and a series of variants~\cite{tsd,rcnet,agvm} such as ATSS~\cite{atss}, RetinaNet~\cite{retina}, FCOS~\cite{fcos}, and PAA~\cite{paa} lead to the significant breakthrough of object detection task. 
One-to-many label assignment is the core scheme of them, where each ground-truth box is assigned to multiple coordinates in the detector's output as the supervised target cooperated with proposals~\cite{fast,faster}, anchors~\cite{retina} or window centers~\cite{fcos}.
Despite their promising performance, these detectors heavily rely on many hand-designed components like a non-maximum suppression procedure or anchor generation~\cite{detr}.
To conduct a more flexible end-to-end detector, DEtection TRansformer (DETR)~\cite{detr} is proposed to view the object detection as a set prediction problem and introduce the one-to-one set matching scheme based on a transformer encoder-decoder architecture.
In this manner, each ground-truth box will only be assigned to one specific query, and multiple hand-designed components that encode prior knowledge are no longer needed.
This approach introduces a flexible detection pipeline and encourages many DETR variants to further improve it.
However, the performance of the vanilla end-to-end object detector is still inferior to the traditional detectors with one-to-many label assignments. 
% Motivated by the success of DETR, many follow-up works have attempted to improve the efficiency and performance of DETR from various aspects, such as \todo{xxx}.
% Even though many end-to-end detectors have been derived based on this one-to-one set matching, their performance is still inferior to traditional detectors with one-to-many label assignment. 
% This naturally raises a question: \emph{can we significately improve the training efficiency and performance of DETR-based detectors and make it superior to conventional detectors while mantaining their end-to-end merit?}

\begin{figure}[t]
    \begin{minipage}[t]{0.48\linewidth}
        \centering
        \includegraphics[width=1\textwidth]{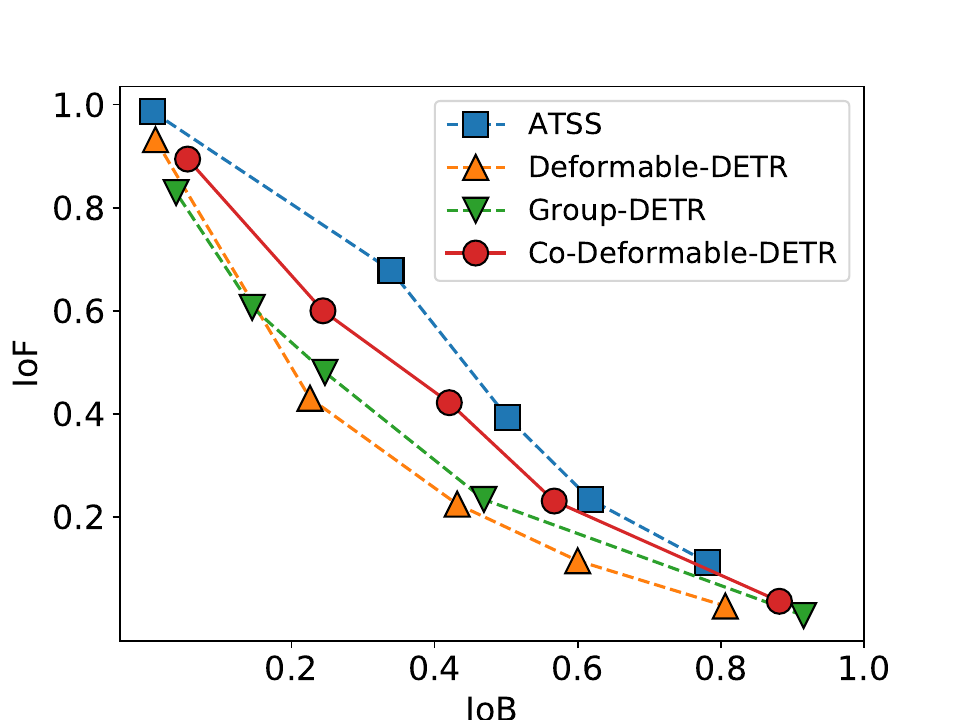}
        \label{encoder_curve}
    \end{minipage}
    % \hspace{1mm}
    \begin{minipage}[t]{0.48\linewidth}
        \centering
        \includegraphics[width=1\textwidth]{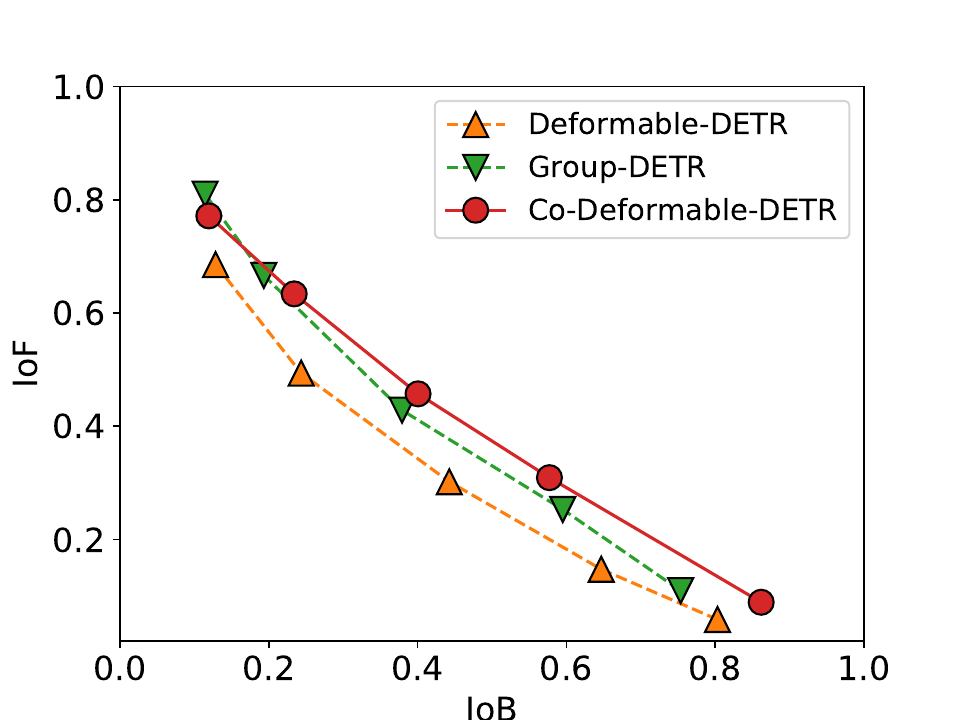}
        \label{decoder_curve}
    \end{minipage}
    \vspace{-0.4cm}
    \caption{
    IoF-IoB curves for the feature discriminability score in the encoder and attention discriminability score in the decoder.
    }
    \vspace{-0.4cm}
    \label{fig:enc_dec}
\end{figure}

\begin{figure}[t]
    \centering
    \includegraphics[width=1\linewidth]{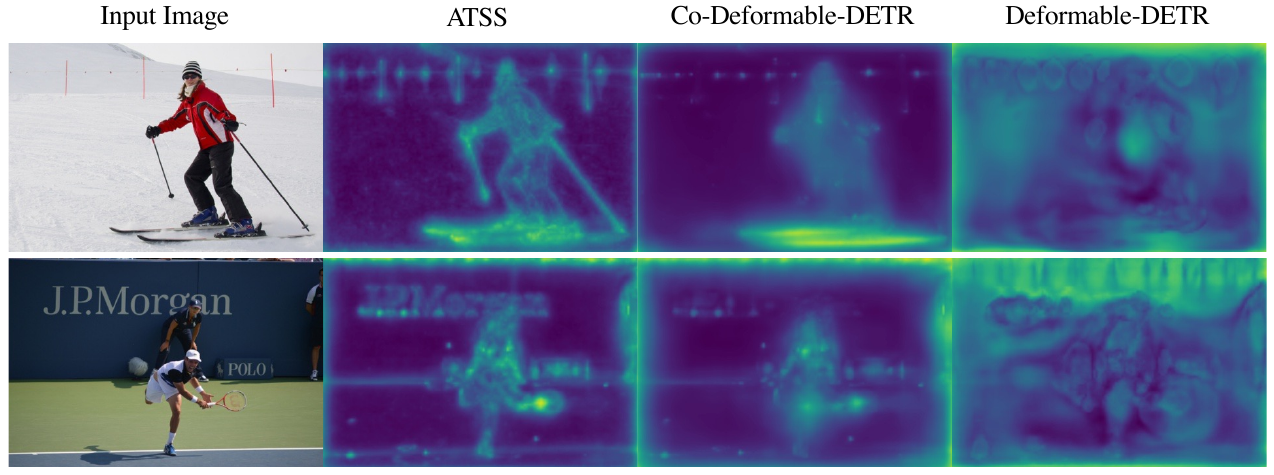}
    \vspace{-0.4cm}
    \caption{Visualizations of discriminability scores in the encoder.}
    \vspace{-0.2cm}
    \label{fig:feature}
\end{figure}

In this paper, we try to make DETR-based detectors superior to conventional detectors while maintaining their end-to-end merit.
To address this challenge, we focus on the intuitive drawback of one-to-one set matching that it explores less positive queries.
This will lead to severe inefficient training issues.
We detailedly analyze this from two aspects, the latent representation generated by the encoder and the attention learning in the decoder.
We first compare the discriminability score of the latent features between the Deformable-DETR~\cite{deformable} and the one-to-many label assignment method where we simply replace the decoder with the ATSS head.
The feature $l^{2}$-norm in each spatial coordinate is utilized to represent the discriminability score.
Given the encoder's output $\mathcal{F}\in \mathbb{R}^{C\times H\times W}$, we can obtain the discriminability score map $\mathcal{S}\in\mathbb{R}^{1\times H\times W}$.
The object can be better detected when the scores in the corresponding area are higher.
As shown in Figure \ref{fig:enc_dec}, we demonstrate the IoF-IoB curve (IoF: intersection over foreground, IoB: intersection over background) by applying different thresholds on the discriminability scores (details in Section~\ref{sec:whywork}).
% Given the latent feature $\mathcal{F}$
% For the computation of IoF and IoB, we define the feature discriminability score as normalized $l^{2}$-norm of feature 
% and set a score thresh $S$.
% Spatial coordinates in the feature map with higher discriminability scores than $S$ activate the foreground while lower ones activate the background.
The higher IoF-IoB curve in ATSS indicates that it's easier to distinguish the foreground and background.
We further visualize the discriminability score map $\mathcal{S}$ in Figure~\ref{fig:feature}.
It's obvious that the features in some salient areas are fully activated in the one-to-many label assignment method but less explored in one-to-one set matching.
For the exploration of decoder training, we also demonstrate the IoF-IoB curve of the cross-attention score in the decoder based on the Deformable-DETR and the Group-DETR~\cite{group} which introduces more positive queries into the decoder.
The illustration in Figure~\ref{fig:enc_dec} shows that too few positive queries also influence attention learning and increasing more positive queries in the decoder can slightly alleviate this. 

This significant observation motivates us to present a simple but effective method, a collaborative hybrid assignment training scheme ($\mathcal{C}$o-DETR).
The key insight of $\mathcal{C}$o-DETR is to use versatile one-to-many label assignments to improve the training efficiency and effectiveness of both the encoder and decoder.
More specifically, we integrate the auxiliary heads with the output of the transformer encoder.
These heads can be supervised by versatile one-to-many label assignments such as ATSS~\cite{atss}, FCOS~\cite{fcos}, and Faster RCNN ~\cite{faster}.
Different label assignments enrich the supervisions on the encoder's output which forces it to be discriminative enough to support the training convergence of these heads.
% To further improve the training efficiency of decoder, we elaborately encode the foreground coordinates in the outputs of these auxiliary heads, including the \todo{}. 
To further improve the training efficiency of the decoder, we elaborately encode the coordinates of positive samples in these auxiliary heads, including the positive anchors and positive proposals. 
They are sent to the original decoder as multiple groups of positive queries to predict the pre-assigned categories and bounding boxes.
Positive coordinates in each auxiliary head serve as an independent group that is isolated from the other groups.
Versatile one-to-many label assignments can introduce lavish (positive query, ground-truth) pairs to improve the decoder's training efficiency.
Note that, only the original decoder is used during inference, thus the proposed training scheme only introduces extra overheads during training.

We conduct extensive experiments to evaluate the efficiency and effectiveness of the proposed method. 
Illustrated in Figure~\ref{fig:feature}, $\mathcal{C}$o-DETR greatly alleviates the poorly encoder's feature learning in one-to-one set matching.
As a plug-and-play approach, we easily combine it with different DETR variants, including DAB-DETR~\cite{dab}, Deformable-DETR~\cite{deformable}, and DINO-Deformable-DETR~\cite{dino}.
As shown in Figure~\ref{fig:performance}, $\mathcal{C}$o-DETR achieves faster training convergence and even higher performance.
Specifically, we improve the basic Deformable-DETR by 5.8\% AP in 12-epoch training and 3.2\% AP in 36-epoch training. 
% The state-of-the-art $\mathcal{H}$-Deformable-DETR with Swin-L~\cite{swin} can still be improved from $57.9$\% to $58.7$\% on MS COCO \texttt{val}.
The state-of-the-art DINO-Deformable-DETR with Swin-L~\cite{swin} can still be improved from 58.5\% to 59.5\% AP on COCO \texttt{val}.
Surprisingly, incorporated with ViT-L~\cite{eva02} backbone, we achieve 66.0\% AP on COCO \texttt{test-dev} and 67.9\% AP on LVIS \texttt{val}, establishing the new state-of-the-art detector with much fewer model sizes.

%% file: 2_related.tex
\section{Related Works}
\label{sec:related}

\begin{figure*}[tp]
    \centering
    \includegraphics[width=\textwidth]{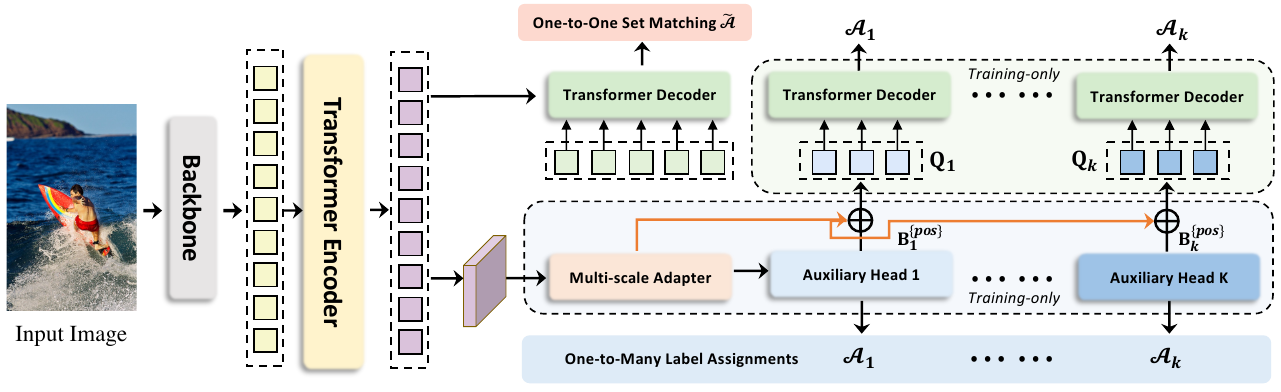}
    \caption{\textbf{Framework of our Collaborative Hybrid Assignment Training.} The auxiliary branches are discarded during evaluation.}
    \label{fig:framework}
    \vspace{-0.2cm}
\end{figure*}

\noindent
\textbf{One-to-many label assignment.}
% Label assignment that determines positive and negative samples is crucial to the optimization of object detector.
% Current label assignment strategies can mainly be categorized as one-to-many assignment and one-to-one assignment.
For one-to-many label assignment in object detection, multiple box candidates can be assigned to the same ground-truth box as positive samples in the training phase.
% Thus, this property contributes to the redundant predictions during inference while bringing numerous positive samples during training.
In classic anchor-based detectors, such as Faster-RCNN \cite{faster} and RetinaNet \cite{retina}, the sample selection is guided by the predefined IoU threshold and matching IoU between anchors and annotated boxes. 
The anchor-free FCOS \cite{fcos} leverages the center priors and assigns spatial locations near the center of each bounding box as positives.
Moreover, the adaptive mechanism is incorporated into one-to-many label assignments to overcome the limitation of fixed label assignments.
ATSS \cite{atss} performs adaptive anchor selection by the statistical dynamic IoU values of top-$k$ closest anchors.
PAA \cite{paa} adaptively separates anchors into positive and negative samples in a probabilistic manner.
In this paper, we propose a collaborative hybrid assignment scheme to improve encoder representations via auxiliary heads with one-to-many label assignments.

\noindent
\textbf{One-to-one set matching.}
The pioneering transformer-based detector, DETR~\cite{detr}, incorporates the one-to-one set matching scheme into object detection and performs fully end-to-end object detection.
The one-to-one set matching strategy first calculates the global matching cost via Hungarian matching and assigns only one positive sample with the minimum matching cost for each ground-truth box.
% DETR abandons the conventional anchor priors while introducing a set of learnable object queries to formulate object detection as a set prediction problem.
% Although DETR achieves comparable performance with Faster-RCNN, the training cost is affordable as it requires more than 300 training epochs to converge.
% Due to the slow convergence issue of DETR, great efforts have been made to accelerate the training process and achieve significant performance.
% Deformable-DETR proposes multi-scale deformable attention to sample features according to several sampling locations of reference points.
% DAB-DETR formulates the object queries as anchors and explicitly uses positional priors to accelerate training.
DN-DETR~\cite{dn} demonstrates the slow convergence results from the instability of one-to-one set matching, thus introducing denoising training to eliminate this issue.
DINO~\cite{dino} inherits the advanced query formulation of DAB-DETR~\cite{dab} and incorporates an improved contrastive denoising technique to achieve state-of-the-art performance.
Group-DETR~\cite{group} constructs group-wise one-to-many label assignment to exploit multiple positive object queries, which is similar to the hybrid matching scheme in $\mathcal{H}$-DETR~\cite{hybrid}.
In contrast with the above follow-up works, we present a new perspective of collaborative optimization for one-to-one set matching.

%% file: 3_method.tex
\section{Method}

\subsection{Overview}
Following the standard DETR protocol, the input image is fed into the backbone and encoder to generate latent features.
Multiple predefined object queries interact with them in the decoder via cross-attention afterwards.
We introduce $\mathcal{C}$o-DETR to improve the feature learning in the encoder and the attention learning in the decoder via the collaborative hybrid assignments training scheme and the customized positive queries generation.
We will detailedly describe these modules and give insights why they can work well.
% To improve the training efficiency, as shown in Figure \ref{fig:framework}, we propose the collaborative hybrid assignment training scheme for DETR ($\mathcal{C}$o-DETR).
% The overall framework of our $\mathcal{C}$o-DETR is simple and can be flexibly adopted by all the DETR variants.
% Specifically, we first incorporate $K$ parallel auxiliary heads after the encoder, such as ATSS and Faster-RCNN, to perform collaborative hybrid training.
% To fully exploit the positive priors from the auxiliary heads, we further leverage their positive priors within the decoder.
% Note that we only perform the collaborative branches in the training phase, thus no extra computation costs are introduced during inference.
\input{label_assign.tex}
\subsection{Collaborative Hybrid Assignments Training}
% \noindent\textbf{$\mathcal{C}$o-DETR.}
To alleviate the sparse supervision on the encoder's output caused by the fewer positive queries in the decoder, 
we incorporate versatile auxiliary heads with different one-to-many label assignment paradigms, \eg, ATSS, and Faster R-CNN.
Different label assignments enrich the supervisions on the encoder’s output which forces it to be discriminative enough
to support the training convergence of these heads.
% Specifically, given the encoder's latent feature $\mathcal{F}$, we firstly transform it to the feature pyramid $\{\mathcal{F}_1, \cdots, \mathcal{F}_J\}$ via the multi-scale adapter where $J\in[1,N]$ indicates feature map with $2^{2+J}$ downsampling stride.
Specifically, given the encoder's latent feature $\mathcal{F}$, we firstly transform it to the feature pyramid $\{\mathcal{F}_1, \cdots, \mathcal{F}_J\}$ via the multi-scale adapter where $J$ indicates feature map with $2^{2+J}$ downsampling stride.
Similar to ViTDet~\cite{vitdet}, the feature pyramid is constructed by a single feature map in the single-scale encoder, while we use bilinear interpolation and $3\times3$ convolution for upsampling.
For instance, with the single-scale feature from the encoder, we successively apply downsampling ($3\times3$ convolution with stride $2$) or upsampling operations to produce a feature pyramid.
% As for multi-scale encoder, latent feature $\mathcal{F}$
% the feature pyramid is naturally constructed, thus we only downsample the coarsest feature to build the feature pyramid.
As for the multi-scale encoder, we only downsample the coarsest feature in the multi-scale encoder features $\mathcal{F}$ to build the feature pyramid.
% This is implemented by ViTDet~\cite{}, where the feature pyramid is built by a single feature map.
Defined $K$ collaborative heads with corresponding label assignment manners $\mathcal{A}_{k}$, for the $i$-th collaborative head, $\{\mathcal{F}_1, \cdots, \mathcal{F}_J\}$ is sent to it to obtain the predictions $\mathbf{\hat{P}}_{i}$.
At the $i$-th head, $\mathcal{A}_{i}$ is used to compute the supervised targets for the positive and negative samples in $\mathbf{P}_{i}$.
Denoted $\mathbf{G}$ as the ground-truth set, this procedure can be formulated as:
\begin{equation}
    \label{eq:assign}
    \mathbf{P}_{i}^{\{pos\}}, \mathbf{B}_{i}^{\{pos\}}, \mathbf{P}_{i}^{\{neg\}} = \mathcal{A}_{i}(\mathbf{\hat{P}}_{i}, \mathbf{G}),
\end{equation}
where ${\{pos\}}$ and ${\{neg\}}$ indicate the pair set of ($j$, positive coordinates or negative coordinates in $\mathcal{F}_j$) determined by $\mathcal{A}_i$.
$j$ means the feature index in $\{\mathcal{F}_1, \cdots, \mathcal{F}_J\}$.
$\mathbf{B}_{i}^{\{pos\}}$ is the set of spatial positive coordinates.
$\mathbf{P}_{i}^{\{pos\}}$ and $\mathbf{P}_{i}^{\{neg\}}$ are the supervised targets in the corresponding coordinates, including the categories and regressed offsets.
To be specific, we describe the detailed information about each variable in Table \ref{tab:label_assign}.
% Specifically, we perform the label assignment procedure $\mathcal{A}^{i}(\cdot)$ for $\mathbf{B}^{i}$ by one-to-many matching between predictions and ground-truth boxes $\mathbf{G}$ as:
% \begin{equation}
%     \label{eq:assign}
%     (\mathbf{B}^{i}_{pos}, \mathbf{G}^{i}_{pos}), (\mathbf{B}^{i}_{neg}, \mathbf{G}^{i}_{neg}) = \mathcal{A}^{i}(\mathbf{B}^{i}, \mathbf{G}),
% \end{equation}
% where $\mathbf{B}^{i}_{pos}$ and $\mathbf{B}^{i}_{neg}$ indicate the positive and negative priors in the $i$-th head, respectively. $\mathbf{G}^{i}$ refers to the corresponding targets. 
% Subsequently, the predictions for priors $\mathbf{B}^{i}_{pos}$ and $\mathbf{B}^{i}_{neg}$, which are denoted by $\mathbf{P}^{i}_{pos}$ and $\mathbf{P}^{i}_{neg}$ participate in the loss calculation of the $i$-th head,
The loss functions can be defined as:
\begin{equation}
    \mathcal{L}^{enc}_{i} = \mathcal{L}_{i}(\mathbf{\hat{P}}_{i}^{\{pos\}}, \mathbf{{P}}_{i}^{\{pos\}}) + \mathcal{L}_{i}(\mathbf{\hat{P}}_{i}^{\{neg\}}, \mathbf{{P}}_{i}^{\{neg\}}),
\end{equation}
% In $\mathcal{L}^{i}$, we adopt the xx loss for classification and xx loss for localization. 
Note that the regression loss is discarded for negative samples.
The training objective of the optimization for $K$ auxiliary heads is formulated as follows:
\begin{equation}
    \mathcal{L}^{enc} = 
    \sum_{i=1}^{K}\mathcal{L}_{i}^{enc}
\end{equation}

\subsection{Customized Positive Queries Generation}

In the one-to-one set matching paradigm, each ground-truth box will only be assigned to one specific query as the supervised target.
Too few positive queries lead to inefficient cross-attention learning in the transformer decoder as shown in Figure \ref{fig:enc_dec}.
To alleviate this, we elaborately generate sufficient customized positive queries according to the label assignment $\mathcal{A}_{i}$ in each auxiliary head.
Specifically,
given the positive coordinates set $\mathbf{B}_{i}^{\{pos\}} \in \mathbb{R}^{M_{i}\times 4}$ in the $i$-th auxiliary head, where $M_{i}$ is the number of positive samples, the extra customized positive queries  $\mathbf{Q}_{i} \in \mathbb{R}^{M_{i}\times C}$ can be generated by:
\begin{equation}
    \mathbf{Q}_{i} = \mathrm{Linear}(\mathrm{PE}(\mathbf{B}_{i}^{\{pos\}}))+\mathrm{Linear}(\mathrm{E}(\mathbf{\{\mathcal{F}_*\}},{\{pos\}})).
\end{equation}
where $\mathrm{PE}(\cdot)$ stands for positional encodings and we select the corresponding features from $\mathrm{E}(\cdot)$ according to the index pair ($j$, positive coordinates or negative coordinates in $\mathcal{F}_j$).

As a result, there are $K+1$ groups of queries that contribute to a single one-to-one set matching branch and $K$ branches with one-to-many label assignments during training.
The auxiliary one-to-many label assignment branches share the same parameters with $L$ decoders layers in the original main branch.
All the queries in the auxiliary branch are regarded as positive queries, thus the matching process is discarded.
To be specific, the loss of the $l$-th decoder layer in the $i$-th auxiliary branch can be formulated as:
\begin{equation}
    \mathcal{L}^{dec}_{i, l} = \widetilde{\mathcal{L}}(\widetilde{\mathbf{P}}_{i, l}, \mathbf{{P}}_{i}^{\{pos\}}).
\end{equation}
$\widetilde{\mathbf{P}}_{i, l}$ refers to the output predictions of the $l$-th decoder layer in the $i$-th auxiliary branch.
% $\mathbf{{P}}^{i}_{\{pos\}}$ stands for the targets assigned by the label assignment of $i$-th collaborative head.
Finally, the training objective for $\mathcal{C}$o-DETR is:
\begin{equation}
    \mathcal{L}^{global} = \sum_{l=1}^{L}(\widetilde{\mathcal{L}}^{dec}_{l} + \lambda_{1}\sum_{i=1}^{K}\mathcal{L}^{dec}_{i, l} + \lambda_{2}\mathcal{L}^{enc}),
\end{equation}
where $\widetilde{\mathcal{L}}^{dec}_{l}$ stands for the loss in the original one-to-one set matching branch~\cite{detr}, $\lambda_{1}$ and $\lambda_{2}$ are the coefficient balancing the losses.

\subsection{Why Co-DETR works}\label{sec:whywork}
$\mathcal{C}$o-DETR leads to evident improvement to the DETR-based detectors. In the following, we try to investigate its effectiveness qualitatively and quantitatively.
We conduct detailed analysis based on Deformable-DETR with ResNet-50~\cite{resnet} backbone using the 36-epoch setting.

\vspace{1mm}
\noindent\textbf{Enrich the encoder's supervisions.}
Intuitively, too few positive queries lead to sparse supervisions as only one query is supervised by regression loss for each ground-truth.
% In Deformable-DETR, xx\% pixels in encoder's output received the gradients from positive samples and  
% by contrast, xx\% pixels in ATSS do this. 
% In Deformable-DETR, $9.38$ pixels per ground-truth in encoder features received stronger supervisions from positive samples and $11.74$ pixels 
The positive samples in one-to-many label assignment manners receive more localization supervisions to help enhance the latent feature learning.
To further explore how the sparse supervisions impede the model training, we detailedly investigate the latent features produced by the encoder.
We introduce the IoF-IoB curve to quantize the discriminability score of the encoder's output.
Specifically, given the latent feature $\mathcal{F}$ of the encoder, inspired by the feature visualization in Figure \ref{fig:feature}, we compute the IoF (intersection over foreground) and IoB (intersection over background).
% We first visualize the feature discriminability of ATSS, Deformable-DETR, and our $\mathcal{C}$o-DETR in Figure.
% Compared with Deformable-DETR, ATSS owns a stronger ability to distinguish the areas of key objects, while Deformable-DETR is almost disturbed by the background.
% Alternatively, the feature visualizations of $C$o-DETR have more concentrated activation maps on foreground than Deformable-DETR.
% Then we set different score thresh for the feature discriminability and plot the curve between intersection over foreground (IoF) and intersection over background (IoB) in Figure, where ATSS generally achieves higher IoF values under the same IoB settings.
% This phenomenon demonstrates the encoder representations benefit from the one-to-many assignment.
Given the encoder's feature $\mathcal{F}_j\in \mathbb{R}^{C\times H_{j} \times W_{j}}$ at level $j$, we first calculate the $l^{2}$-norm $\widehat{\mathcal{F}}_{j}\in \mathbb{R}^{1\times H_{j} \times W_{j}}$ and resize it to the image size $H \times W$. The discriminability score $\mathcal{D}(\mathcal{F})$ is computed by averaging the scores from all levels:
\begin{equation}
    \mathcal{D}(\mathcal{F}) = \frac{1}{J}\sum_{j=1}^{J}\frac{\widehat{\mathcal{F}}_{j}}{max(\widehat{\mathcal{F}}_{j})},
\end{equation}
where the resize operation is omitted.
% which can be viewed as normalization.
We visualize the discriminability scores of ATSS, Deformable-DETR, and our $\mathcal{C}$o-Deformable-DETR in Figure \ref{fig:feature}.
Compared with Deformable-DETR, both ATSS and $\mathcal{C}$o-Deformable-DETR own stronger ability to distinguish the areas of key objects, while Deformable-DETR is almost disturbed by the background. 
Consequently, we define the indicators for foreground and background as $\mathbbm{1}(\mathcal{D}(\mathcal{F})>S) \in \mathbb{R}^{H \times W}$ and $\mathbbm{1}(\mathcal{D}(\mathcal{F})<S)\in \mathbb{R}^{H \times W}$, respectively.
% \begin{equation}
%     I^{j}_{enc} = \mathbbm{1}(\mathcal{D}_{enc}(\mathcal{F}_j)>S),
% \end{equation}
% where $S_{enc}$ is a predefined score thresh, $\mathbbm{1}(x)$ is 1 if $x$ is true and 0 otherwise.
$S$ is a predefined score thresh, $\mathbbm{1}(x)$ is 1 if $x$ is true and 0 otherwise.
As for the mask of foreground $\mathcal{M}^{fg} \in \mathbb{R}^{H \times W}$, the element $\mathcal{M}^{fg}_{h,w}$ is 1 if the point $(h, w)$ is inside the foreground and 0 otherwise.
The area of intersection over foreground (IoF) $\mathcal{I}^{fg}$ can be computed as:
\begin{equation}
    \mathcal{I}^{fg} = \frac{\sum_{h=1}^{H}\sum_{w=1}^{W}(\mathbbm{1}(\mathcal{D}(\mathcal{F}_{h,w})>S) \cdot \mathcal{M}^{fg}_{h,w})}{\sum_{h=1}^{H}\sum_{w=1}^{W}\mathcal{M}^{fg}_{h,w}}.
\end{equation}
Concretely, we compute the area of intersection over background areas (IoB) in a similar way and plot the curve IoF and IoB by varying $S$ in Figure \ref{fig:enc_dec}.
Obviously, ATSS and $C$o-Deformable-DETR obtain higher IoF values than both Deformable-DETR and Group-DETR under the same IoB values, which demonstrates the encoder representations benefit from the one-to-many label assignment.

% \begin{figure}[t]
%     \begin{minipage}[t]{0.46\linewidth}
%         \centering
%         \includegraphics[width=1\textwidth]{fig/adapter.pdf}
%         \label{adapter1}
%     \end{minipage}
%     \hspace{1mm}
%     \begin{minipage}[t]{0.49\linewidth}
%         \centering
%         \includegraphics[width=1\textwidth]{fig/adapter2.pdf}
%         \label{adapter2}
%     \end{minipage}
%     \vspace{-0.5cm}
%     \caption{
%     Our multi-scale adapter. The left one is applicable to the single-scale encoder and another for multi-scale encoder.
%     }
%     \vspace{-0.3cm}
%     \label{pic:adapter}
% \end{figure}

\begin{figure}[t] 
    \centering
    \includegraphics[width=0.85\linewidth]{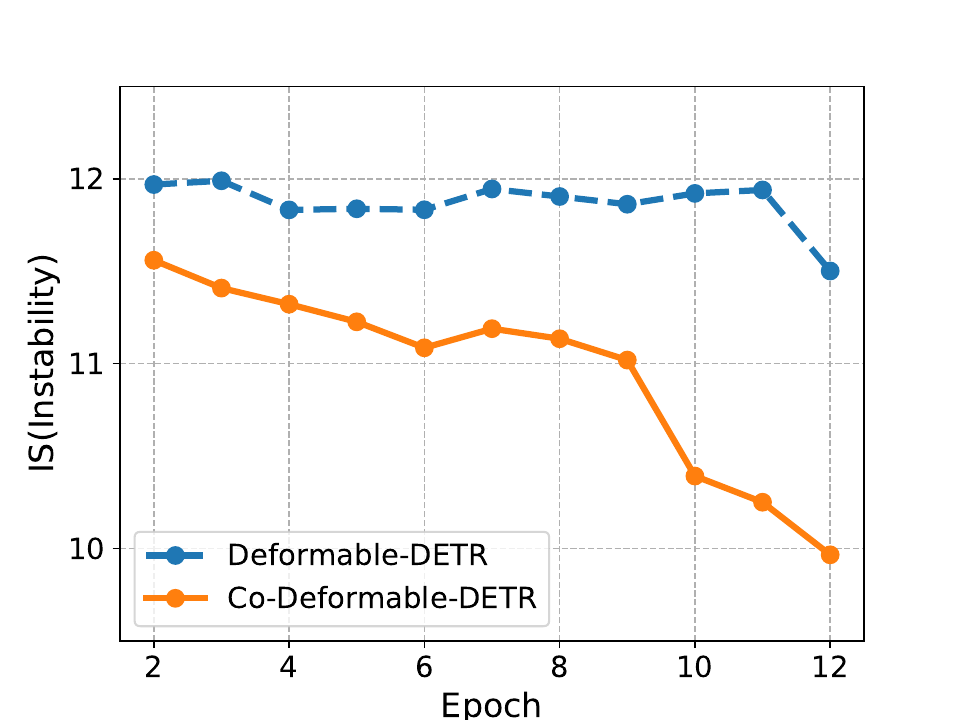}
    \vspace{-0.2cm}
    \caption{The instability (IS)~\cite{dn} of Deformable-DETR and $\mathcal{C}$o-Deformable-DETR on COCO dataset. These detectors are trained for 12 epochs with ResNet-50 backbones.}
    \label{fig:instability}
    % \vspace{-2mm}
\end{figure}
 
\vspace{1mm}
\noindent\textbf{Improve the cross-attention learning by reducing the instability of Hungarian matching.}
Hungarian matching is the core scheme in one-to-one set matching.
Cross-attention is an important operation to help the positive queries encode abundant object information.
It requires sufficient training to achieve this.
We observe that the Hungarian matching introduces uncontrollable  instability since the ground-truth assigned to a specific positive query in the same image is changing during the training process.
Following~\cite{dn}, we present the comparison of instability in Figure \ref{fig:instability}, where we find our approach contributes to a more stable matching process.
Furthermore, in order to quantify how well cross-attention is being optimized, we also calculate the IoF-IoB curve for attention score.
Similar to the feature discriminability score computation, we set different thresholds for attention score to get multiple IoF-IoB pairs.
The comparisons between Deformable-DETR, Group-DETR, and $\mathcal{C}$o-Deformable-DETR can be viewed in Figure \ref{fig:enc_dec}.
We find that the IoF-IoB curves of DETRs with more positive queries are generally above Deformable-DETR, which is consistent with our motivation.

\subsection{Comparison with other methods}
\noindent\textbf{Differences between our method and other counterparts.}
Group-DETR, $\mathcal{H}$-DETR, and SQR~\cite{sqr} perform one-to-many assignments by one-to-one matching with duplicate groups and repeated ground-truth boxes.
$\mathcal{C}$o-DETR explicitly assigns multiple spatial coordinates as positives for each ground truth.
Accordingly, these dense supervision signals are directly applied to the latent feature map to enable it more discriminative.
By contrast, Group-DETR, $\mathcal{H}$-DETR, and SQR lack this mechanism.
% More positive queries are introduced in Group-DETR and $\mathcal{H}$-DETR to facilitate the decoder attention learning, where the feature learning is omitted.
% Thanks to the collaborative training, we enable discriminative feature and attention learning to achieve superior performance to these counterparts (see Table \textcolor{red}{9} and L\textcolor{red}{800}).
Although more positive queries are introduced in these counterparts, the one-to-many assignments implemented by Hungarian Matching still suffer from the instability issues of one-to-one matching. 
Our method benefits from the stability of off-the-shelf one-to-many assignments and inherits their specific matching manner between positive queries and ground-truth boxes.
Group-DETR and $\mathcal{H}$-DETR fail to reveal the complementarities between one-to-one matching and traditional one-to-many assignment.
To our best knowledge, we are the first to give the quantitative and qualitative analysis on the detectors with the traditional one-to-many assignment and one-to-one matching. This helps us better understand their differences and complementarities so that we can naturally improve the DETR's learning ability by leveraging off-the-shelf one-to-many assignment designs without requiring additional specialized one-to-many design experience.

\noindent\textbf{No negative queries are introduced in the decoder.}
Duplicate object queries inevitably bring large amounts of negative queries for the decoder and a significant increase in GPU memory.
However, our method only processes the positive coordinates in the decoder, thus consuming less memory as shown in Table \ref{tab:multi_head}.

%% file: label_assign.tex
\begin{table*}[tp]
    \centering
    \setlength\tabcolsep{2pt}
    \resizebox{\textwidth}{!}{
    \begin{tabular}{l|c|c|c|c}
        \Xhline{1.0pt}
        \multirow{2}*{Head $i$} & \multirow{2}*{Loss $\mathcal{L}_{i}$} & \multicolumn{3}{c}{Assignment $\mathcal{A}_{i}$} \\
        \cline{3-5}
        ~ & ~ & ${\{pos\}}$, ${\{neg\}}$ Generation &  $\mathbf{P}_{i}$ Generation & $\mathbf{B}_{i}^{\{pos\}}$ Generation \\
        % Head $i$ & Assignment $\mathcal{A}^{i}$ & ${\{pos\}}$ Generation & ${\{neg\}}$ Generation & $\mathbf{B}^{i}_{\{pos\}}$ Generation \\
        \hline
        \multirow{2}*{Faster-RCNN~\cite{faster}} & cls: CE loss, & $\{pos\}$: IoU(proposal, gt)\textgreater $0.5$ & $\{pos\}$: gt labels, offset(proposal, gt) & positive proposals \\
         & reg: GIoU loss & $\{neg\}$: IoU(proposal, gt)\textless $0.5$& $\{neg\}$: gt labels & $(x_{1}, y_{1}, x_{2}, y_{2})$ \\
        \hline
        \multirow{2}*{ATSS~\cite{atss}} & cls: Focal loss & $\{pos\}$:IoU(anchor, gt)\textgreater (mean+std) & $\{pos\}$: gt labels, offset(anchor, gt), centerness &  positive anchors \\
         & reg: GIoU, BCE loss & $\{neg\}$: IoU(anchor, gt)\textless(mean+std) & $\{neg\}$: gt labels & $(x_{1}, y_{1}, x_{2}, y_{2})$ \\
        \hline
        \multirow{2}*{RetinaNet~\cite{retina}} & cls: Focal loss & $\{pos\}$: IoU(anchor, gt)\textgreater$0.5$ & $\{pos\}$: gt labels, offset(anchor, gt) & positive anchors \\
         & reg: GIoU Loss & $\{neg\}$: IoU(anchor, gt)\textless$0.4$ & $\{neg\}$: gt labels & $(x_{1}, y_{1}, x_{2}, y_{2})$ \\ 
        \hline
        \multirow{2}*{FCOS~\cite{fcos}} & cls: Focal Loss & $\{pos\}$: points inside gt center area & $\{pos\}$: gt labels, ltrb distance, centerness & FCOS point $(cx, cy)$ \\
         & reg: GIoU, BCE loss & $\{neg\}$: points outside gt center area & $\{neg\}$: gt labels & $w=h=8\times2^{2+j}$ \\
        \Xhline{1.0pt}
    \end{tabular}
    }
    \vspace{-0.2cm}
    \caption{\textbf{Detailed information of auxiliary heads.} The auxiliary heads include Faster-RCNN~\cite{faster}, ATSS~\cite{atss}, RetinaNet~\cite{retina}, and FCOS~\cite{fcos}. If not otherwise specified, we follow the original implementations, \eg, anchor generation.}
    \vspace{-0.2cm}
    \label{tab:label_assign}
\end{table*}

%% file: 4_results.tex
\section{Experiments}

\input{plain_results.tex}

\subsection{Setup}

\vspace{1mm}
\noindent\textbf{Datasets and Evaluation Metrics.}
Our experiments are conducted on the MS COCO 2017 dataset~\cite{coco} and LVIS v1.0 dataset~\cite{lvis}. The COCO dataset consists of 115K labeled images for training and 5K images for validation. 
We report the detection results by default on the \texttt{val} subset. 
The results of our largest model evaluated on the \texttt{test-dev} (20K images) are also reported. 
LVIS v1.0 is a large-scale and long-tail dataset with 1203 categories for large vocabulary instance segmentation.
To verify the scalability of $\mathcal{C}$o-DETR, we further apply it to a large-scale object detection benchmark, namely Objects365~\cite{objects365}. 
There are 1.7M labeled images used for training and 80K images for validation in the Objects365 dataset. 
All results follow the standard mean Average Precision(AP) under IoU thresholds ranging from 0.5 to 0.95 at different object scales.

\vspace{1mm}
\noindent\textbf{Implementation Details.}
We incorporate our $\mathcal{C}$o-DETR into the current DETR-like pipelines and keep the training setting consistent with the baselines.
We adopt ATSS and Faster-RCNN as the auxiliary heads for $K=2$ and only keep ATSS for $K=1$.
More details about our auxiliary heads can be found in the supplementary materials. 
% Following the DETR protocol, we use the AdamW optimizer with learning rate $\frac{batch\ size}{16} \times 10^{-4}$ for transformer, $\frac{batch\ size}{16} \times 10^{-5}$ for backbone, and $10^{-4}$ as weight decay.
% The learning rate is multiplied by a factor of 0.1 at the $11$-th and $30$-th for the scheduler of $12$ and $36$ epochs.
We choose the number of learnable object queries to 300 and set $\{\lambda_{1}, \lambda_{2}\}$ to $\{1.0, 2.0\}$ by default.
For $\mathcal{C}$o-DINO-Deformable-DETR++, we use large-scale jitter with copy-paste~\cite{copypaste}.
% No more than 300 positive anchor queries per head are randomly sampled for the auxiliary positive prior learning.
% We choose the number of co-trained heads $K$ to 1 by default. 
% For image augmentation during training, we perform the multi-scale training and random crop augmentation adopted in DETR.
% In our case, the loss of decoder is composed of classificiation loss~\cite{retina}, $l_{1}$ loss and GIoU loss~\cite{giou} that is consistent with the original implementation.

\subsection{Main Results}
In this section, we empirically analyze the effectiveness and generalization ability of $\mathcal{C}$o-DETR on different DETR variants in Table~\ref{tab:plain_results} and Table~\ref{tab:main_results}.
All results are reproduced using mmdetection~\cite{mmdet}.
% Both the results of $K=1$ and $K=2$ are reported in Table \ref{tab:main_results} and Table \ref{tab:backbone_results}.
% \vspace{1mm}
% \noindent\textbf{Results with ResNet-50.}
We first apply the collaborative hybrid assignments training to single-scale DETRs with C5 features.
Surprisingly, both Conditional-DETR and DAB-DETR obtain 2.4\% and 2.3\% AP gains over the baselines with a long training schedule.
For Deformable-DETR with multi-scale features, the detection performance is significantly boosted from 37.1\% to 42.9\% AP.
The overall improvements (+3.2\% AP) still hold when the training time is increased to 36 epochs.
Moreover, we conduct experiments on the improved Deformable-DETR (denoted as Deformable-DETR++) following~\cite{hybrid}, where a +2.4\% AP gain is observed.
% The state-of-the-art $\mathcal{H}$-Deformable-DETR equipped with our method can achieve $49.7\%$, which is $+1.3\%$ higher than the competitive baseline. 
The state-of-the-art DINO-Deformable-DETR equipped with our method can achieve 51.2\% AP, which is +1.8\% AP higher than the competitive baseline. 
% As for $K=1$, we also observe our method consistently yields significant gains over the baselines and the performance is marginally lower compared with $K=2$.

We further scale up the backbone capacity from ResNet-50 to Swin-L~\cite{swin} based on two state-of-the-art baselines.
As presented in Table \ref{tab:main_results}, $\mathcal{C}$o-DETR achieves 56.9\% AP and surpasses the Deformable-DETR++ baseline by a large margin (+1.7\% AP).
The performance of DINO-Deformable-DETR with Swin-L can still be boosted from 58.5\% to 59.5\% AP.

\input{main_results.tex}
\input{sota_coco.tex}

\input{sota.tex}
\subsection{Comparisons with the state-of-the-art}
We apply our method with $K=2$ to Deformable-DETR++ and DINO.
Besides, the quality focal loss~\cite{gfl} and NMS are adopted for our $\mathcal{C}$o-DINO-Deformable-DETR.
We report the comparisons on COCO \texttt{val} in Table \ref{tab:coco_sota}.
% Note that the results of $K=1$ are released in the supplementary materials.
Compared with other competitive counterparts, our method converges much faster.
For example, $\mathcal{C}$o-DINO-Deformable-DETR readily achieves 52.1\% AP when using only 12 epochs with ResNet-50 backbone.
Our method with Swin-L can obtain 58.9\% AP for 1$\times$ scheduler, even surpassing other state-of-the-art frameworks on 3$\times$ scheduler.
More importantly, our best model $\mathcal{C}$o-DINO-Deformable-DETR++ achieves 54.8\% AP with ResNet-50 and 60.7\% AP with Swin-L under 36-epoch training, outperforming all existing detectors with the same backbone by clear margins.
% We further use the Swin Transformer as the backbone and increase the training time to a longer schedule of 36 epochs.
% We can see that our models consistently outperform the state-of-the-art $\mathcal{H}$-Deformable-DETR with the same backbone by $\sim1.0\%$ AP.
% More importantly, $\mathcal{H}$-Deformable-DETR can still be improved from $57.9\%$ to $58.7\%$, surpassing the state-of-the-art DINO-Deformable-DETR ($58.0\%$) that equipped with an improved query formulation and denoising training.
% More importantly, the performance of DINO-Deformable-DETR can still be boosted from $58.5\%$ to $59.5\%$, surpassing other state-of-the-art methods by a large margin.
% our best model with Swin-L reaches $\mathbf{58.5\%}$, surpassing the state-of-the-art DINO-Deformable-DETR ($58.0\%$) that equipped with a improved query formulation and denoising training.

To further explore the scalability of our method, we extend the backbone capacity to 304 million parameters.
This large-scale backbone ViT-L~\cite{vit} is pre-trained using a self-supervised learning method (EVA-02~\cite{eva02}).
% compare with previous state-of-the-art detection frameworks pre-trained with additional data.
% This extremely large backbone MixMIM-g~\cite{mixmim} is pre-trained on ImageNet-1K dataset~\cite{imagenet} using self-supervised learning.
% More details about MixMIM-g are provided in the supplementary materials.
% We first pre-train our model with MixMIM-g as the backbone on Objects$365$ using 32 A100 GPUs.
% As shown in Table \ref{tab:obj365}, our model establishes a new record $51.5\%$, which is $+12.2\%$ higher than the previous state-of-the-art Florence, while requiring less than $99.8\%$ pre-training data.
% We further fine-tune the pre-trained detector on COCO dataset for 20 epochs.
We first pre-train $\mathcal{C}$o-DINO-Deformable-DETR with ViT-L on Objects365 for 26 epochs, then fine-tune it on the COCO dataset for 12 epochs.
In the fine-tuning stage, the input resolution is randomly selected between 480$\times$2400 and 1536$\times$2400.
The detailed settings are available in supplementary materials.
Our results are evaluated with test-time augmentation. 
Table \ref{tab:sota} presents the state-of-the-art comparisons on the COCO \texttt{test-dev} benchmark.
With much fewer model sizes (304M parameters), $\mathcal{C}$o-DETR sets a new record of 66.0\% AP on COCO \texttt{test-dev}, outperforming the previous best model InternImage-G~\cite{intern} by +0.5\% AP.

We also demonstrate the best results of $\mathcal{C}$o-DETR on the long-tailed LVIS detection dataset.
In particular, we use the same $\mathcal{C}$o-DINO-Deformable-DETR++ as the model on COCO but choose FedLoss~\cite{centernet2} as the classification loss to remedy the impact of unbalanced data distribution.
Here, we only apply bounding boxes supervision and report the object detection results.
The comparisons are available in Table~\ref{tab:sota_lvis}.
$\mathcal{C}$o-DETR with Swin-L yields 56.9\% and 62.3\% AP on LVIS \texttt{val} and \texttt{minival}, surpassing ViTDet with MAE-pretrained~\cite{mae} ViT-H and GLIPv2~\cite{glipv2} by +3.5\% and +2.5\% AP, respectively.
% which is +3.5\% points higher than ViTDet with MAE-pretrained~\cite{mae} ViT-H. 
We further finetune the Objects365 pretrained $\mathcal{C}$o-DETR on this dataset.
Without elaborate test-time augmentation, our approach achieves the best detection performance of 67.9\% and 71.9\% AP on LVIS \texttt{val} and \texttt{minival}.
Compared to the 3-billion parameter InternImage-G with test-time augmentation, we obtain +4.7\% and +6.1\% AP gains on LVIS \texttt{val} and \texttt{minival} while reducing the model size to 1/10.

\input{sota_lvis.tex}

\subsection{Ablation Studies}
Unless stated otherwise, all experiments for ablations are conducted on Deformable-DETR with a ResNet-50 backbone.
We choose the number of auxiliary heads $K$ to 1 by default and set the total batch size to 32.
More ablations and analyses can be found in the supplementary materials.

\vspace{1mm}
\noindent\textbf{Criteria for choosing auxiliary heads.}
% To further delve into the influence of $K$, we report the performance and training efficiency by controlling the number of auxiliary heads.
% In Table \ref{tab:multi_head}, we find $\mathcal{C}$o-DETR with ATSS performs better than hybrid matching scheme~\cite{hybrid} while training faster and requiring less GPU memory.
% It is worth noting that the accuracy continues to increase as the training costs of the model increase when choosing $K$ smaller than 3.
% We speculate the optimization conflicts among these heads lead to performance degradation when $K \geq 4$.
% Overall, we choose $K \leq 2$ and Faster-RCNN as the second auxiliary head since it achieves the best trade-offs between training efficiency and accuracy.
% To further delve into the combination of different heads, we report the performance by controlling the number of auxiliary heads.
We further delve into the criteria for choosing auxiliary heads in Table \ref{tab:multi_head} and \ref{tab:heads_ablation}.
The results in Table \ref{tab:heads_ablation} reveal that \textit{any} auxiliary head with one-to-many label assignments consistently improves the baseline and ATSS achieves the best performance. 
We find the accuracy continues to increase as $K$ increases when choosing $K$ smaller than 3.
It is worth noting that performance degradation occurs when $K=6$, and we speculate the severe conflicts among auxiliary heads cause this.
If the feature learning is inconsistent across the auxiliary heads, the continuous improvement as $K$ becomes larger will be destroyed.
% For example, PAA cannot bring gains and even hurts the performance when cooperating with other heads.
We also analyze the optimization consistency of multiple heads next and in the supplementary materials.
In summary, we can choose any head as the auxiliary head and we regard ATSS and Faster-RCNN as the common practice to achieve the best performance when $K\leq2$.
We do not use too many different heads, \eg, 6 different heads to avoid optimization conflicts.

\input{multi_head.tex}
\input{dense_head.tex}

\begin{figure}[t] 
    \centering
    \includegraphics[width=0.9\linewidth]{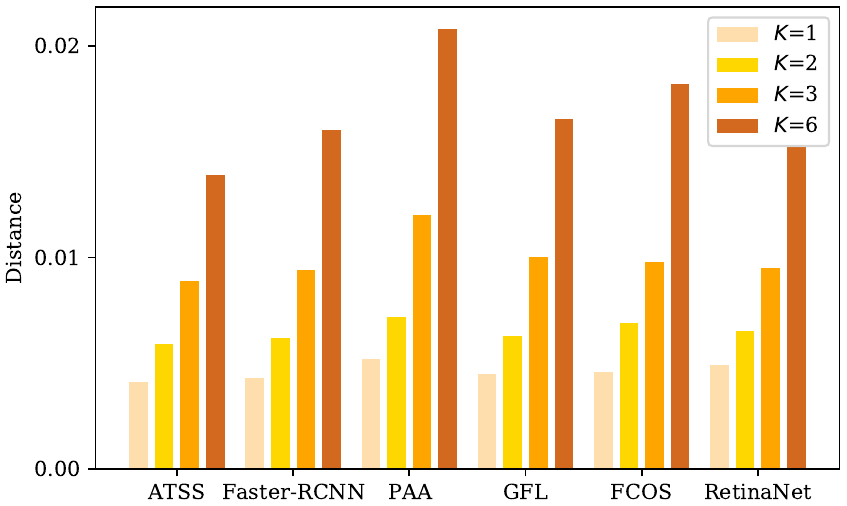}
    \vspace{-0.1cm}
    \caption{The distance when varying $K$ from 1 to 6.}
    \label{fig:cam}
\end{figure}

\vspace{1mm}
\noindent\textbf{Conflicts analysis.} 
The conflicts emerge when the same spatial coordinate is assigned to different foreground boxes or treated as background in different auxiliary heads and can confuse the training of the detector. 
We first define the distance between head $H_i$ and head $H_j$, and the average distance of $H_i$ to measure the optimization conflicts as:
\begin{equation}
    \mathcal{S}_{i,j} =\frac{1}{|\mathbf{D}|} \sum_{\mathbf{I} \in \mathbf{D}} \mathrm{KL}(\mathcal{C}(H_i(\mathbf{I})), \mathcal{C}(H_j(\mathbf{I})),
\end{equation}
\begin{equation}
    \mathcal{S}_{i} =\frac{1}{2(K-1)} \sum_{j \neq i}^{K} (\mathcal{S}_{i,j}+\mathcal{S}_{j,i}),
\end{equation}
where $\mathrm{KL}$, $\mathbf{D}$, $\mathbf{I}$, $\mathcal{C}$ refer to KL divergence, dataset, the input image, and class activation maps (CAM)~\cite{cam}. 
As illustrated in Figure~\ref{fig:cam}, we compute the average distances among auxiliary heads for $K>1$ and the distance between the DETR head and the single auxiliary head for $K=1$.
We find the distance metric is \textit{insignificant} for each auxiliary head when $K=1$ and this observation is consistent with our results in Table~\ref{tab:heads_ablation}: the DETR head can be collaboratively improved with any head when $K=1$.
When $K$ is increased to 2, the distance metrics increase \textit{slightly} and our method achieves the best performance as shown in Table~\ref{tab:multi_head}.
The distance \textit{surges} when $K$ is increased from 3 and 6, indicating severe optimization conflicts among these auxiliary heads lead to a decrease in performance. 
However, the baseline with 6 ATSS achieves 49.5\% AP and can be decreased to 48.9\% AP by replacing ATSS with 6 various heads. 
Accordingly, we speculate \textit{too many} diverse auxiliary heads, \eg, more than 3 different heads, exacerbate the conflicts.
In summary, optimization conflicts are influenced by the number of various auxiliary heads and the relations among these heads.

\input{component.tex}

\vspace{1mm}
\noindent \textbf{Should the added heads be different?}
Collaborative training with two ATSS heads (49.2\% AP) still improves the model with one ATSS head (48.7\% AP) as ATSS is complementary to the DETR head in our analysis.
% As presented in Figure \textcolor{red}{9} and our experiments, the optimization target of ATSS is consistent with the DETR head, thus more ATSS heads for our method further improve the performance.
Besides, introducing a diverse and complementary auxiliary head rather than the same one as the original head, \eg, Faster-RCNN, can bring better gains (49.5\% AP).
Note that this is \textit{not contradictory} to above conclusion; instead, we can obtain the best performance with \textit{few different heads} ($K\leq2$) as the conflicts are insignificant, but we are faced with severe conflicts when using \textit{many different heads} ($K>3$).

\vspace{1mm}
\noindent\textbf{The effect of each component.}
We perform a component-wise ablation to thoroughly analyze the effect of each component in Table \ref{tab:component}. 
Incorporating the auxiliary head yields significant gains since the dense spatial supervision enables the encoder features more discriminative.
% Simply incorporating a auxiliary head for the encoder optimization yields $+4.5$ AP and $+2.9$ AP gains under $12$ and $36$ training epochs, respectively.
% We also see introducing the positive samples assigned by one-to-many matching into the decoder improves the baseline from $37.1$ AP to $40.5$ AP.
Alternatively, introducing customized positive queries also contributes remarkably to the final results, while improving the training efficiency of the one-to-one set matching.
Both techniques can accelerate convergence and improve performance.
In summary, we observe the overall improvements stem from more discriminative features for the encoder and more efficient attention learning for the decoder.

\input{longer_schedule.tex}
\input{individual.tex}

\vspace{1mm}
\noindent\textbf{Comparisons to the longer training schedule.}
As presented in Table \ref{tab:longer_schedule}, we find Deformable-DETR can not benefit from longer training as the performance saturates.
On the contrary, $\mathcal{C}$o-DETR greatly accelerates the convergence as well as increasing the peak performance.

\vspace{1mm}
\noindent\textbf{Performance of auxiliary branches.}
Surprisingly, we observe $\mathcal{C}$o-DETR also brings consistent gains for auxiliary heads in Table \ref{tab:cotrain}.
This implies our training paradigm contributes to more discriminative encoder representations, which improves the performances of both decoder and auxiliary heads.

\begin{figure}[t]
    \begin{minipage}[t]{0.45\linewidth}
        \centering
        \includegraphics[width=1\textwidth]{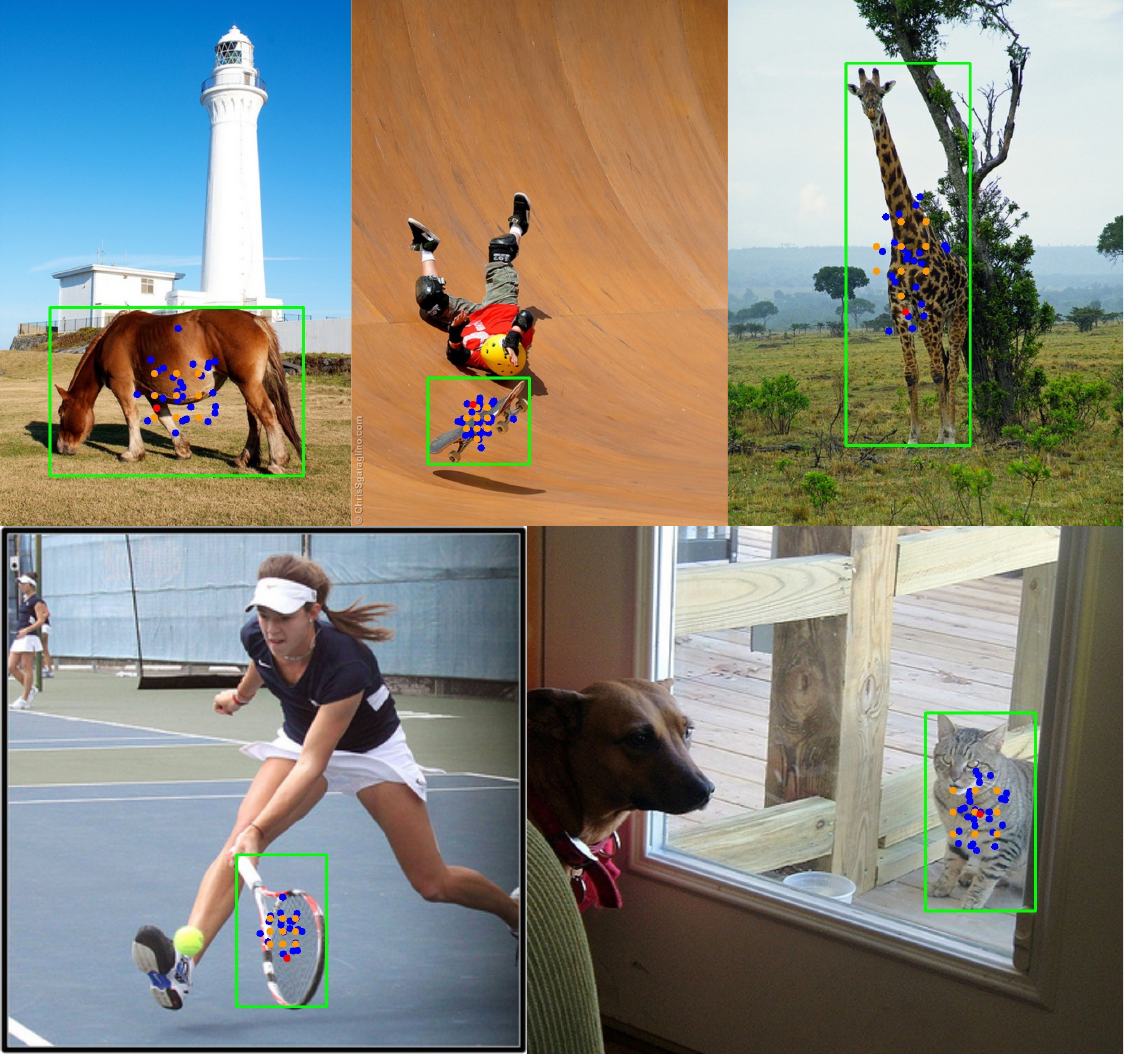}
        \subcaption{Visualizations of queries.}
        \label{fig:query}
    \end{minipage}
    \hspace{2mm}
    \begin{minipage}[t]{0.45\linewidth}
        \centering
        \includegraphics[width=1\textwidth]{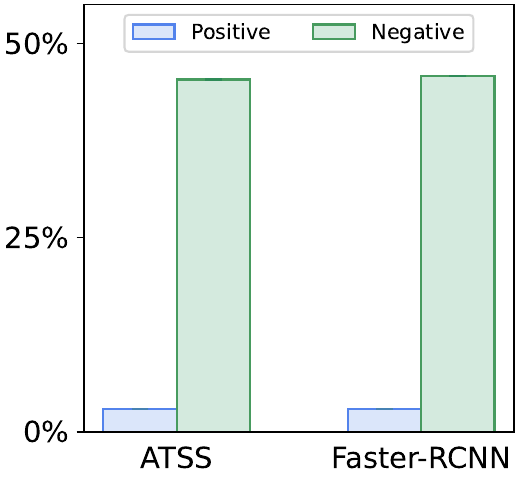}
        \subcaption{Normalized distances.}
        \label{fig:distance}
    \end{minipage}
    \label{fig:distribution}
    \vspace{-0.2cm}
    \caption{
    Distribution of original and customized queries.}
    \vspace{-0.1cm}
\end{figure}

\vspace{1mm}
\noindent\textbf{Difference in distribution of original and customized positive queries.}
We visualize the positions of original positive queries and customized positive queries in Figure~\ref{fig:query}.
We only show one object (green box) per image. 
Positive queries assigned by Hungarian Matching in the decoder are marked in red. 
We mark positive queries extracted from Faster-RCNN and ATSS in blue and orange, respectively.
These customized queries are distributed around the center region of the instance and provide sufficient supervision signals for the detector.

\vspace{1mm}
\noindent\textbf{Does distribution difference lead to instability?}
We compute the average distance between original and customized queries in Figure~\ref{fig:distance}.
The average distance between original negative queries and customized positive queries is significantly larger than the distance between original and customized positive queries.
As this distribution gap between original and customized queries is marginal, there is no instability encountered during training.

%% file: plain_results.tex
\begin{table}[t]
    \centering\setlength{\tabcolsep}{6pt}
    \footnotesize
    \renewcommand{\arraystretch}{1.2}
    \resizebox{0.95\linewidth}{!}
    {
        \begin{tabular}{l|c|c|c}
        \shline
        Method & $K$ & \#epochs  & AP \\
        \shline
        Conditional DETR-C5~\cite{conditional} & 0 & 36 & 39.4 \\ 
        Conditional DETR-C5~\cite{conditional} & 1 & 36 & 41.5\color[RGB]{17, 122, 101}{\textbf{(+2.1)}} \\ 
        Conditional DETR-C5~\cite{conditional} & 2 & 36 & 41.8\color[RGB]{17, 122, 101}{\textbf{(+2.4)}} \\ 
        \hline
        DAB-DETR-C5~\cite{dab} & 0 & 36 & 41.2 \\ 
        DAB-DETR-C5~\cite{dab} & 1 & 36 & 43.1\color[RGB]{17, 122, 101}{\textbf{(+1.9)}} \\ 
        DAB-DETR-C5~\cite{dab} & 2 & 36 & 43.5\color[RGB]{17, 122, 101}{\textbf{(+2.3)}} \\ 
        \hline
        Deformable-DETR~\cite{deformable} & 0 & 12 & 37.1 \\ 
        Deformable-DETR~\cite{deformable} & 1 & 12 & 42.3\color[RGB]{17, 122, 101}{\textbf{(+5.2)}} \\ 
        Deformable-DETR~\cite{deformable} & 2 & 12 & 42.9\color[RGB]{17, 122, 101}{\textbf{(+5.8)}} \\ 
        \hline
        Deformable-DETR~\cite{deformable} & 0 & 36 & 43.3 \\ 
        Deformable-DETR~\cite{deformable} & 1 & 36 & 46.8\color[RGB]{17, 122, 101}{\textbf{(+3.5)}} \\ 
        Deformable-DETR~\cite{deformable} & 2 & 36 & 46.5\color[RGB]{17, 122, 101}{\textbf{(+3.2)}} \\ 
        \shline
        \end{tabular}
    }
    \vspace{-2mm}
    \caption{\small{Results of plain baselines on COCO \texttt{val}.}
    }
    \label{tab:plain_results}
    % \vspace{-5mm}
\end{table}

%% file: main_results.tex
\begin{table}[t]
    \centering\setlength{\tabcolsep}{6pt}
    \footnotesize
    \renewcommand{\arraystretch}{1.2}
    \resizebox{\linewidth}{!}
    {
        \begin{tabular}{l|c|c|c}
        \shline
        Method & $K$ & \#epochs  & AP \\
        \shline
        % Conditional DETR-C5~\cite{conditional} & 0 & 36 & 39.4 \\ 
        % Conditional DETR-C5~\cite{conditional} & 1 & 36 & 41.5\color[RGB]{17, 122, 101}{\textbf{(+2.1)}} \\ 
        % Conditional DETR-C5~\cite{conditional} & 2 & 36 & 41.8\color[RGB]{17, 122, 101}{\textbf{(+2.4)}} \\ 
        % \hline
        % DAB-DETR-C5~\cite{dab} & 0 & 36 & 41.2 \\ 
        % DAB-DETR-C5~\cite{dab} & 1 & 36 & 43.1\color[RGB]{17, 122, 101}{\textbf{(+1.9)}} \\ 
        % DAB-DETR-C5~\cite{dab} & 2 & 36 & 43.5\color[RGB]{17, 122, 101}{\textbf{(+2.3)}} \\ 
        % \hline
        % Deformable-DETR~\cite{deformable} & 0 & 12 & 37.1 \\ 
        % Deformable-DETR~\cite{deformable} & 1 & 12 & 42.3\color[RGB]{17, 122, 101}{\textbf{(+5.2)}} \\ 
        % Deformable-DETR~\cite{deformable} & 2 & 12 & 42.9\color[RGB]{17, 122, 101}{\textbf{(+5.8)}} \\ 
        % \hline
        % Deformable-DETR~\cite{deformable} & 0 & 36 & 43.3 \\ 
        % Deformable-DETR~\cite{deformable} & 1 & 36 & 46.8\color[RGB]{17, 122, 101}{\textbf{(+3.5)}} \\ 
        % Deformable-DETR~\cite{deformable} & 2 & 36 & 46.5\color[RGB]{17, 122, 101}{\textbf{(+3.2)}} \\ 
        % \hline
        Deformable-DETR++~\cite{deformable} & 0 & 12 & 47.1 \\ 
        Deformable-DETR++~\cite{deformable} & 1 & 12 & 48.7\color[RGB]{17, 122, 101}{\textbf{(+1.6)}} \\
        Deformable-DETR++~\cite{deformable} & 2 & 12 & 49.5\color[RGB]{17, 122, 101}{\textbf{(+2.4)}} \\
        % \hline
        % $\mathcal{H}$-Deformable-DETR~\cite{hybrid} & 0 & 12 & 48.4 \\ 
        % $\mathcal{H}$-Deformable-DETR~\cite{hybrid} & 1 & 12 & 49.2\color[RGB]{17, 122, 101}{\textbf{(+0.8)}} \\ 
        % $\mathcal{H}$-Deformable-DETR~\cite{hybrid} & 2 & 12 & 49.7\color[RGB]{17, 122, 101}{\textbf{(+1.3)}} \\ 
        \hline
        DINO-Deformable-DETR{$^\dag$}~\cite{dino} & 0 & 12 & 49.4  \\ 
        DINO-Deformable-DETR{$^\dag$}~\cite{dino} & 1 & 12 & 51.0\color[RGB]{17, 122, 101}{\textbf{(+1.6)}} \\ 
        DINO-Deformable-DETR{$^\dag$}~\cite{dino} & 2 & 12 & 51.2\color[RGB]{17, 122, 101}{\textbf{(+1.8)}} \\ 
        \hline 
        Deformable-DETR++{$^\ddag$}~\cite{deformable} & 0 & 12 & 55.2 \\ 
        Deformable-DETR++{$^\ddag$}~\cite{deformable} & 1 & 12 & 56.4\color[RGB]{17, 122, 101}{\textbf{(+1.2)}} \\
        Deformable-DETR++{$^\ddag$}~\cite{deformable} & 2 & 12 & 56.9\color[RGB]{17, 122, 101}{\textbf{(+1.7)}} \\
        \hline
        DINO-Deformable-DETR{$^\dag$}{$^\ddag$}~\cite{dino} & 0 & 36 & 58.5 \\ 
        DINO-Deformable-DETR{$^\dag$}{$^\ddag$}~\cite{dino} & 1 & 36 & 59.3\color[RGB]{17, 122, 101}{\textbf{(+0.8)}} \\
        DINO-Deformable-DETR{$^\dag$}{$^\ddag$}~\cite{dino} & 2 & 36 & 59.5\color[RGB]{17, 122, 101}{\textbf{(+1.0)}} \\ 
        \shline
        \end{tabular}
    }
    \vspace{-2mm}
    \caption{\small{Results of strong baselines on COCO \texttt{val}. Methods with $\dag$ use 5 feature levels. $\ddag$ refers to Swin-L backbone.}
    }
    \label{tab:main_results}
    % \vspace{-5mm}
\end{table}

%% file: sota_coco.tex
\begin{table*}[htbp]
    \centering
    \setlength{\tabcolsep}{6pt}
    \footnotesize
    \renewcommand{\arraystretch}{1.2}
    \resizebox{\linewidth}{!}
    {
        \begin{tabular}{l|l|c|c|c|cccccc}
        \shline
        Method  & Backbone & Multi-scale & \#query & \#epochs  & AP & AP$_{50}$ & AP$_{75}$ & AP$_{S}$ & AP$_{M}$ & AP$_{L}$ \\
        \shline
        Conditional-DETR~\cite{conditional} & R50 & \xmark & 300 & 108 & 43.0 & 64.0 & 45.7 & 22.7 & 46.7 & 61.5 \\
        Anchor-DETR~\cite{anchordetr} & R50 & \xmark & 300 & 50 & 42.1 & 63.1 & 44.9 & 22.3 & 46.2 & 60.0 \\
        DAB-DETR~\cite{dab} & R50 & \xmark & 900 & 50 & 45.7 & 66.2 & 49.0 & 26.1 & 49.4 & 63.1 \\
        AdaMixer~\cite{adamixer} & R50 & \cmark & 300 & 36 & 47.0 & 66.0 & 51.1 & 30.1 & 50.2 & 61.8 \\
        % AdaMixer~\cite{adamixer} & R101 & \cmark & 300 & 36 & 48.0 & 67.0 & 52.4 & 30.0 & 51.2 & 63.7 \\
        % AdaMixer~\cite{adamixer} & Swin-S & \cmark & 300 & 36 & 51.3 & 71.2 & 55.7 & 34.2 & 54.6 & 67.3 \\
        Deformable-DETR~\cite{deformable} & R50 & \cmark & 300 & 50 & 46.9 & 65.6 & 51.0 & 29.6 & 50.1 & 61.6 \\
        DN-Deformable-DETR~\cite{dn} & R50 & \cmark & 300 & 50 & 48.6 & 67.4 & 52.7 & 31.0 & 52.0 & 63.7 \\
        DINO-Deformable-DETR{$^\dag$}~\cite{dino} & R50 & \cmark & 900 & 12 & 49.4 & 66.9 & 53.8 & 32.3 & 52.5 & 63.9 \\ 
        DINO-Deformable-DETR{$^\dag$}~\cite{dino} & R50 & \cmark & 900 & 36 & 51.2 & 69.0 & 55.8 & 35.0 & 54.3 & 65.3 \\ 
        DINO-Deformable-DETR{$^\dag$}~\cite{dino} & Swin-L (IN-22K) & \cmark & 900 & 36 & 58.5 & 77.0 & 64.1 & 41.5 & 62.3 & 74.0 \\ 
        Group-DINO-Deformable-DETR~\cite{group} & Swin-L (IN-22K) & \cmark & 900 & 36 & 58.4 & - & - & 41.0 & 62.5 & 73.9 \\
        $\mathcal{H}$-Deformable-DETR~\cite{hybrid} & R50 & \cmark & 300 & 12 & 48.7 & 66.4 & 52.9 & 31.2 & 51.5 & 63.5 \\
        % $\mathcal{H}$-Deformable-DETR~\cite{hybrid} & R101 & \cmark & 300 & 12 & 49.4 & 67.2 & 53.7 & 31.9 & 53.1 & 64.2 \\
        % $\mathcal{H}$-Deformable-DETR~\cite{hybrid} & Swin-T & \cmark & 300 & 36 & 53.2 & 71.5 & 58.2 & 35.9 & 56.4 & 68.2 \\
        % $\mathcal{H}$-Deformable-DETR~\cite{hybrid} & Swin-S & \cmark & 300 & 36 & 54.4 & 72.9 & 59.4 & 36.9 & 58.3 & 69.5 \\
        % $\mathcal{H}$-Deformable-DETR & Swin-L & \cmark & 300 & 12 & 55.9 & 75.2 & 61.0 & 39.1 & 59.9 & 72.2 \\
        % $\mathcal{H}$-Deformable-DETR~\cite{hybrid} & Swin-L (IN-$22$K) & \cmark & 300 & 36 & $57.1$ & $76.2$ & $62.5$ & $39.7$ & $61.4$ & $73.4$ \\
        $\mathcal{H}$-Deformable-DETR~\cite{hybrid} & Swin-L (IN-22K) & \cmark & 900 & 36 & 57.9 & 76.8 & 63.6 & 42.4 & 61.9 & 73.4 \\
        \hline
        \rowcolor[gray]{.9}
        $\mathcal{C}$o-Deformable-DETR & R50 & \cmark & 300 & 12 & 49.5 & 67.6 & 54.3 & 32.4 & 52.7 & 63.7 \\
        \rowcolor[gray]{.9}
        $\mathcal{C}$o-Deformable-DETR & Swin-L (IN-22K) & \cmark & 900 & 36 & 58.5 & 77.1 & 64.5 & 42.4 & 62.4 & 74.0 \\
        \hline
        % $\mathcal{C}$o-Deformable-DETR & R101 & \cmark & 300 & 12 & 50.1 & 67.9 & 54.9 & 33.0 & 53.9 & 64.6 \\
        % $\mathcal{C}$o-Deformable-DETR & Swin-T & \cmark & 300 & 12 & $51.7$ & $69.7$ & $56.8$ & $35.3$ & $55.2$ & $67.0$  \\
        % $\mathcal{C}$o-Deformable-DETR & Swin-S & \cmark & 300 & 12 & $53.4$ & $71.8$ & $58.5$ & $35.4$ & $57.0$ & $69.0$  \\
        % $\mathcal{C}$o-Deformable-DETR & Swin-B (IN-$22$K) & \cmark & 300 & 12 & $55.5$ & $74.1$ & $60.7$ & $37.9$ & $59.6$ & $72.2$  \\
        % $\mathcal{C}$o-Deformable-DETR & Swin-L (IN-$22$K) & \cmark & 300 & 12 & $56.9$ & $75.5$ & $62.6$ & $40.1$ & $61.2$ & $73.3$  \\
        \rowcolor[gray]{.9}
        $\mathcal{C}$o-DINO-Deformable-DETR{$^\dag$} & R50 & \cmark & 900 & 12 & 52.1 & 69.4 & 57.1 & 35.4 & 55.4 & 65.9 \\
        % $\mathcal{C}$o-Deformable-DETR & Swin-T & \cmark & 300 & 36 & $54.1$ & $72.4$ & $59.0$ & $37.8$ & $57.3$ & $68.7$ \\
        % $\mathcal{C}$o-Deformable-DETR & Swin-S & \cmark & 300 & 36 & 55.3 & 73.6 & 60.9 & 39.0 & 59.0 & 70.1 \\
        % $\mathcal{C}$o-Deformable-DETR & Swin-B (IN-$22$K) & \cmark & 300 & 36 & $57.5$ & $76.0$ & $63.4$ & $41.3$ & $61.7$ & $73.2$ \\
        % $\mathcal{C}$o-Deformable-DETR & Swin-L (IN-$22$K) & \cmark & 300 & 36 & $58.1$ & $76.6$ & $63.7$ & $40.8$ & $62.2$ & $73.7$  \\
        % $\mathcal{C}$o-Deformable-DETR & Swin-L (IN-$22$K) & \cmark & 900 & 36 & $58.3$ & $76.7$ & $64.3$ & $42.1$ & $62.2$ & $73.9$ \\
        % \rowcolor[gray]{.9}
        % $\mathcal{C}$o-$\mathcal{H}$-Deformable-DETR & Swin-L (IN-$22$K) & \cmark & 900 & 36 & $58.7}$ & $77.4}$ & $64.5}$ & $41.9}$ & $62.7}$ & $74.9}$ \\
        \rowcolor[gray]{.9}
        $\mathcal{C}$o-DINO-Deformable-DETR{$^\dag$} & Swin-L (IN-22K) & \cmark & 900 & 12 & 58.9 & 76.9 & 64.8 & 42.6 & 62.7 & 75.1 \\
        \rowcolor[gray]{.9}
        $\mathcal{C}$o-DINO-Deformable-DETR{$^\dag$} & Swin-L (IN-22K) & \cmark & 900 & 24 & 59.8 & 77.7 & 65.5 & 43.6 & 63.5 & 75.5 \\
        \rowcolor[gray]{.9}
        $\mathcal{C}$o-DINO-Deformable-DETR{$^\dag$} & Swin-L (IN-22K) & \cmark & 900 & 36 & 60.0 & 77.7 & 66.1 & 44.6 & 63.9 & 75.7 \\
        \hline
        \rowcolor[gray]{.9}
        $\mathcal{C}$o-DINO-Deformable-DETR++{$^\dag$} & R50 & \cmark & 900 & 12 & \textbf{52.1} & 69.3 & 57.3 & 35.4 & 55.5 & 67.2 \\
        \rowcolor[gray]{.9}
        $\mathcal{C}$o-DINO-Deformable-DETR++{$^\dag$} & R50 & \cmark & 900 & 36 & \textbf{54.8} & 72.5 & 60.1 & 38.3 & 58.4 & 69.6 \\
        \rowcolor[gray]{.9}
        $\mathcal{C}$o-DINO-Deformable-DETR++{$^\dag$} & Swin-L (IN-22K) & \cmark & 900 & 12 & \textbf{59.3} & 77.3 & 64.9 & 43.3 & 63.3 & 75.5 \\
        \rowcolor[gray]{.9}
        $\mathcal{C}$o-DINO-Deformable-DETR++{$^\dag$} & Swin-L (IN-22K) & \cmark & 900 & 24 & \textbf{60.4} & 78.3 & 66.4 & 44.6 & 64.2 & 76.5 \\
        \rowcolor[gray]{.9}
        $\mathcal{C}$o-DINO-Deformable-DETR++{$^\dag$} & Swin-L (IN-22K) & \cmark & 900 & 36 & \textbf{60.7} & 78.5 & 66.7 & 45.1 & 64.7 & 76.4 \\
        \shline
        \multicolumn{11}{l}{$\dag$: 5 feature levels.}
        \end{tabular}
    }
    \vspace{-3mm}
    \caption{\small{Comparison to the state-of-the-art DETR variants on COCO \texttt{val}.}
    }
    \label{tab:coco_sota}
    \vspace{-0.1cm}
\end{table*}

%% file: sota.tex
\begin{table}[t]
    \centering\setlength{\tabcolsep}{6pt}
    \footnotesize
    \renewcommand{\arraystretch}{1.3}
    \resizebox{\linewidth}{!}
    {
        \begin{tabular}{l|c|c|c|c}
        \shline
        \multirow{2}*{Method} & \multirow{2}*{Backbone} & enc. & \texttt{val} & \texttt{test-dev} \\
         & & \#params & $\text{AP}^{box}$ & $\text{AP}^{box}$ \\
        \shline
         HTC++~\cite{htc} & SwinV2-G~\cite{swinv2} & 3.0B & 62.5 & 63.1 \\
         DINO~\cite{dino} & Swin-L~\cite{swin} & 218M & 63.2 & 63.3 \\
         BEIT3~\cite{beit3} & ViT-g~\cite{vit} & 1.9B & - & 63.7 \\
         FD~\cite{fd} & SwinV2-G~\cite{swinv2} & 3.0B & - & 64.2 \\
         DINO~\cite{dino} & FocalNet-H~\cite{focalnet} & 746M & 64.2 & 64.3 \\
         Group DETRv2~\cite{groupv2} & ViT-H~\cite{vit} & 629M & - & 64.5 \\
         EVA-02~\cite{eva02} & ViT-L~\cite{vit} & 304M & 64.1 & 64.5 \\
         DINO~\cite{dino} & InternImage-G~\cite{intern} & 3.0B & 65.3 & 65.5 \\
         % DINO~\cite{dino} & InternImage-H~\cite{intern} & 2.18B & Laion-400M + YFCC-15M + CC12M + IN-22K (14M) & Objects365 & 65.0 & 65.4 \\
        \hline
        \rowcolor[gray]{.9}
        $\mathcal{C}$o-DETR & ViT-L~\cite{vit} & \textbf{304M} & \textbf{65.9} & \textbf{66.0} \\
        \shline
        \end{tabular}
    }
    \vspace{-2mm}
    \caption{Comparison to the state-of-the-art frameworks on COCO.}
    \vspace{-0.3cm}
    \label{tab:sota}
\end{table}

%% file: sota_lvis.tex
\begin{table}[t]
    \centering\setlength{\tabcolsep}{6pt}
    \footnotesize
    \renewcommand{\arraystretch}{1.3}
    \resizebox{\linewidth}{!}
    {
        \begin{tabular}{l|c|c|c|c}
        \shline
        \multirow{2}*{Method} & \multirow{2}*{Backbone} & enc. & \texttt{val} & \texttt{minival} \\
         & & \#params & $\text{AP}^{box}$ & $\text{AP}^{box}$ \\
        \shline
         $\mathcal{H}$-DETR~\cite{hybrid} & Swin-L~\cite{swin} & 218M & 47.9 & - \\
         ViTDet~\cite{vitdet} & ViT-L~\cite{vit} & 307M & 51.2 & - \\
         ViTDet~\cite{vitdet} & ViT-H~\cite{vit} & 632M & 53.4 & - \\
         GLIPv2~\cite{glipv2} & Swin-H~\cite{swin} & 637M & - & 59.8 \\
         DINO~\cite{dino} & InternImage-G~\cite{intern} & 3.0B & 63.2 & 65.8 \\
         EVA-02~\cite{eva02} & ViT-L~\cite{vit} & 304M & 65.2 & - \\
         % DINO~\cite{dino} & InternImage-H~\cite{intern} & 2.18B & Laion-400M + YFCC-15M + CC12M + IN-22K (14M) & Objects365 & 65.0 & 65.4 \\
        \hline
        \rowcolor[gray]{.9}
        $\mathcal{C}$o-DETR & Swin-L~\cite{swin} & 218M & 56.9 & 62.3 \\
        \rowcolor[gray]{.9}
        $\mathcal{C}$o-DETR & ViT-L~\cite{vit} & \textbf{304M} & \textbf{67.9} & \textbf{71.9} \\
        \shline
        \end{tabular}
    }
    \vspace{-2mm}
    \caption{Comparison to the state-of-the-art frameworks on LVIS.}
    \vspace{-0.2cm}
    \label{tab:sota_lvis}
\end{table}

%% file: multi_head.tex
\begin{table}[t]
    \centering\setlength{\tabcolsep}{6pt}
    \footnotesize
    \renewcommand{\arraystretch}{1.3}
    \resizebox{\linewidth}{!}
    {
        \begin{tabular}{l|c|c|c|c|c}
        \shline
        \multirow{2}{*}{Method} & \multirow{2}{*}{$K$} & Auxiliary & Memory & GPU & \multirow{2}{*}{AP}  \\
        & & head & (MB) & hours & \\         
        \shline
        Deformable-DETR++ & 0 & - & 12808 & 70 & 47.1 \\ 
        \hline
        $\mathcal{H}$-Deformable-DETR & 0 & - & 15307 & 104 & 48.4 \\
        \hline
        \cellcolor{gray!20}Deformable-DETR++ & \cellcolor{gray!20}1 & \cellcolor{gray!20}ATSS & \cellcolor{gray!20}13947 & \cellcolor{gray!20}86 & \cellcolor{gray!20}48.7 \\ 
        \hline
        Deformable-DETR++ & 2 & ATSS + PAA & 14629 & 124 & 49.0 \\ 
        \hline
        \cellcolor{gray!20}Deformable-DETR++ & \cellcolor{gray!20}2 & \cellcolor{gray!20}ATSS + Faster-RCNN & \cellcolor{gray!20}14387 & \cellcolor{gray!20}120 & \cellcolor{gray!20}\textbf{49.5} \\ 
        \hline
        \multirow{2}{*}{Deformable-DETR++} & \multirow{2}{*}{3} & ATSS + Faster-RCNN & \multirow{2}{*}{15263} & \multirow{2}{*}{150} & \multirow{2}{*}{\textbf{49.5}} \\ 
         & & + PAA & & & \\
        % \hline
        % \multirow{2}{*}{Deformable-DETR++} & \multirow{2}{*}{$4$} & ATSS + Faster-RCNN & \multirow{2}{*}{$17144$} & \multirow{2}{*}{$198$} & \multirow{2}{*}{$49.1$} \\ 
        %  & & + PAA + GFL & & & \\
         \hline
        \multirow{3}{*}{Deformable-DETR++} & \multirow{3}{*}{6} & ATSS + Faster-RCNN & \multirow{3}{*}{19385} & \multirow{3}{*}{280} & \multirow{3}{*}{48.9} \\ 
         & & + PAA + RetinaNet & & & \\
         & & + FCOS + GFL & & & \\
        % \hline
        % Deformable-DETR & ATSS & - & - & $36$ & $46.8$ & $65.1$ & $51.5$ \\
        %\rowcolor{blue!15}Deformable-DETR + ATSS & $36$ & $\underline{46.8}$ \\
        \shline
        \end{tabular}
    }
    \vspace{-2mm}
    \caption{\small{Experimental results of $K$ varying from 1 to 6.}
    }
    \label{tab:multi_head}
\end{table}

%% file: dense_head.tex
\begin{table}[t]
    \centering\setlength{\tabcolsep}{6pt}
    \footnotesize
    \renewcommand{\arraystretch}{1.2}
    \resizebox{0.9\linewidth}{!}
    {
        \begin{tabular}{l|c|c|c|c}
        \shline
        Auxiliary head  & \#epochs  & AP & AP$_{50}$ & AP$_{75}$ \\
        \shline
        Baseline & 36 & 43.3 & 62.3 & 47.1 \\ 
        \hline
        RetinaNet~\cite{retina} & 36 & 46.1 & 64.2 & 50.1 \\
        Faster-RCNN~\cite{faster} & 36 & 46.3 & 64.7 & 50.5 \\
        Mask-RCNN~\cite{mask} & 36 & 46.5 & 65.0 & 50.6 \\
        FCOS~\cite{fcos} & 36 & 46.5 & 64.8 & 50.7 \\
        PAA~\cite{paa} & 36 & 46.5 & 64.6 & 50.7 \\
        GFL~\cite{gfl} & 36 & 46.5 & 65.0 & 51.0 \\
        %\rowcolor{blue!15}Deformable-DETR + ATSS & 36 & \underline{46.8} \\
        \cellcolor{gray!20}ATSS~\cite{atss} & \cellcolor{gray!20}36 & \cellcolor{gray!20}\textbf{46.8} & \cellcolor{gray!20}\textbf{65.1} & \cellcolor{gray!20}\textbf{51.5} \\
        \shline
        \end{tabular}
    }
    \vspace{-2mm}
    \caption{\small{Performance of our approach with various auxiliary one-to-many heads on COCO \texttt{val}.}
    }
    \label{tab:heads_ablation}
    \vspace{-3mm}
\end{table}

%% file: component.tex
\begin{table}[t]
    \centering
    \footnotesize
    \renewcommand{\arraystretch}{1.2}
    \resizebox{1.0\linewidth}{!}
    {
        \begin{tabular}{c|c|c|c|c|c}
        \shline
        aux head & pos queries & \#epochs & AP & AP$_{50}$ & AP$_{75}$ \\
        \shline
        \multirow{2}*{\xmark}  & \multirow{2}*{\xmark} & 12 & 37.1 & 55.5 & 40.0 \\
         & & 36 & 43.3 & 62.3 & 47.1 \\
        \hline
        \multirow{2}*{\cmark}  & \multirow{2}*{\xmark} & 12 & 41.6\color[RGB]{17, 122, 101}{\textbf{(+4.5)}} & 59.8 & 45.6 \\
         & & 36 & 46.2\color[RGB]{17, 122, 101}{\textbf{(+2.9)}} & 64.7 & 50.9 \\
        \hline
        \multirow{2}*{\xmark}  & \multirow{2}*{\cmark} & 12 & 40.5\color[RGB]{17, 122, 101}{\textbf{(+3.4)}} & 58.8 & 44.4 \\
         & & 36 & 45.3\color[RGB]{17, 122, 101}{\textbf{(+2.0)}} & 63.5 & 49.8 \\
        \hline
        \multirow{2}*{\cmark}  & \multirow{2}*{\cmark} & 12 & 42.3\color[RGB]{17, 122, 101}{\textbf{(+5.2)}} & 60.5 & 46.1 \\
         & & 36 & 46.8\color[RGB]{17, 122, 101}{\textbf{(+3.5)}} & 65.1 & 51.5 \\
        \shline
        \end{tabular}
    }
    \vspace{-3mm}
    \caption{``aux head'' denotes training with an auxiliary head and ``pos queries'' means the customized positive queries generation.
    }
    \label{tab:component}
    \vspace{-2mm}
\end{table}

%% file: longer_schedule.tex
\begin{table}[t]
    \centering\setlength{\tabcolsep}{6pt}
    \footnotesize
    \renewcommand{\arraystretch}{1.2}
    \resizebox{1.0\linewidth}{!}
    {
        \begin{tabular}{l|c|c|c|c}
        \shline
        % \multirow{2}{*}{Method} & \multirow{2}{*}{$K$} & \multirow{2}{*}{Epochs} & Training & \multirow{2}{*}{AP} \\
        Method & $K$ & \#epochs & GPU hours & AP \\
         % & & & time \\
        \shline
        Deformable-DETR & 1 & 36 & 288 & 46.8 \\ 
        \hline
        Deformable-DETR & 0 & 50 & 333 & 44.5 \\
        Deformable-DETR & 0 & 100 & 667 & 46.0 \\ 
        Deformable-DETR & 0 & 150 & 1000 & 45.9 \\ 
        \shline
        \end{tabular}
        % 0.275 0.485 0.596
    }
    \vspace{-2mm}
    \caption{Comparison to baselines with longer schedule.}
    \label{tab:longer_schedule}
    % \vspace{-2mm}
\end{table}

%% file: individual.tex
\begin{table}[t]
    \centering\setlength{\tabcolsep}{6pt}
    \footnotesize
    \renewcommand{\arraystretch}{1.2}
    \resizebox{1.0\linewidth}{!}
    {
        \begin{tabular}{l|c|c|c|c}
        \shline
        Branch & NMS & $K=0$ & $K=1$ & $K=2$ \\
        \shline
        Deformable-DETR++ & \xmark & 47.1 & 48.7\color[RGB]{17, 122, 101}{\textbf{(+1.6)}} & 49.5\color[RGB]{17, 122, 101}{\textbf{(+2.4)}} \\
        ATSS & \cmark & 46.8 & 47.4\color[RGB]{17, 122, 101}{\textbf{(+0.6)}} & 48.0\color[RGB]{17, 122, 101}{\textbf{(+1.2)}} \\
        Faster-RCNN & \cmark & 45.9 & - & 46.7\color[RGB]{17, 122, 101}{\textbf{(+0.8)}} \\
        \shline
        \end{tabular}
    }
    \vspace{-2mm}
    \caption{Collaborative training consistently improves performances of all branches on Deformable-DETR++ with ResNet-50.}
    \vspace{-1mm}
    \label{tab:cotrain}
\end{table}
% atss, k3, 12e, 0.480 0.666 0.528 0.321 0.515 0.612
% atss, k2, 12e, 0.474 0.658 0.519 0.300 0.513 0.606
% atss, k0, 12e, 0.468 0.651 0.514 0.310 0.508 0.595
% faster, k0, 12e, 0.459 0.666 0.501 0.285 0.500 0.599
% faster, k3, 12e, 0.467 0.675 0.510 0.304 0.505 0.601

%% file: 5_conclusions.tex
\section{Conclusions}
In this paper,
we present a novel collaborative hybrid assignments training scheme, namely $\mathcal{C}$o-DETR, to learn more efficient and effective DETR-based detectors from versatile label assignment manners.
This new training scheme can easily enhance the encoder's learning ability in end-to-end detectors by training the multiple parallel auxiliary heads supervised by one-to-many label assignments.
In addition, we conduct extra customized positive queries by extracting the positive coordinates from these auxiliary heads to improve the training efficiency of positive samples in decoder.
Extensive experiments on COCO dataset demonstrate the efficiency and effectiveness of $\mathcal{C}$o-DETR.
Surprisingly, incorporated with ViT-L backbone, we achieve 66.0\% AP on COCO \texttt{test-dev} and 67.9\% AP on LVIS \texttt{val}, establishing the new state-of-the-art detector with much fewer model sizes.

% \vspace{1mm}
% \noindent\textbf{Acknowledgments.}
% We thank all.

%% file: X_supplementary.tex
\appendix
% --- PDF will be split by an editor (e.g. macOS preview), so need to restart from page 1
% \setcounter{page}{1}

% --- repeat the title (AT: haven't found a more elegant way to do this...)
\twocolumn[
\centering
\Large
\textbf{DETRs with Collaborative Hybrid Assignments Training} \\
\vspace{0.5em}Supplementary Material \\
\vspace{1.0em}
] %< twocolumn

\section{More ablation studies}
\input{num_conv.tex}
\input{loss_head.tex}
\vspace{1mm}
\noindent\textbf{The number of stacked convolutions.}
Table \ref{tab:num_conv} reveals our method is robust for the number of stacked convolutions in the auxiliary head (trained for 12 epochs).
Concretely, we simply choose only 1 shared convolution to enable lightweight while achieving higher performance.

\vspace{1mm}
\noindent\textbf{Loss weights of collaborative training.}
Experimental results related to weighting the coefficient $\lambda_{1}$ and $\lambda_{2}$ are presented in Table \ref{tab:loss_head}. 
We find the proposed method is quite insensitive to the variations of $\{\lambda_{1}, \lambda_{2}\}$, since the performance slightly fluctuates when varying the loss coefficients.
In summary, the coefficients $\{\lambda_{1}, \lambda_{2}\}$ are robust and we set $\{\lambda_{1}, \lambda_{2}\}$ to $\{1.0, 2.0\}$ by default.

\begin{figure}[t] 
    \centering
    \includegraphics[width=0.85\linewidth]{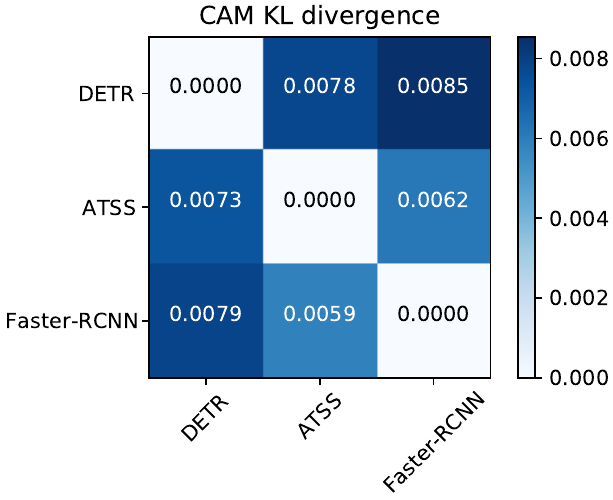}
    \vspace{-0.2cm}
    \caption{The relation matrix for the DETR head, ATSS head, and Faster-RCNN head. The detector is $\mathcal{C}$o-Deformable-DETR ($K=2$) with ResNet-50.}
    \label{fig:mat_k2}
\end{figure}

\begin{figure}[t] 
    \centering
    \includegraphics[width=0.98\linewidth]{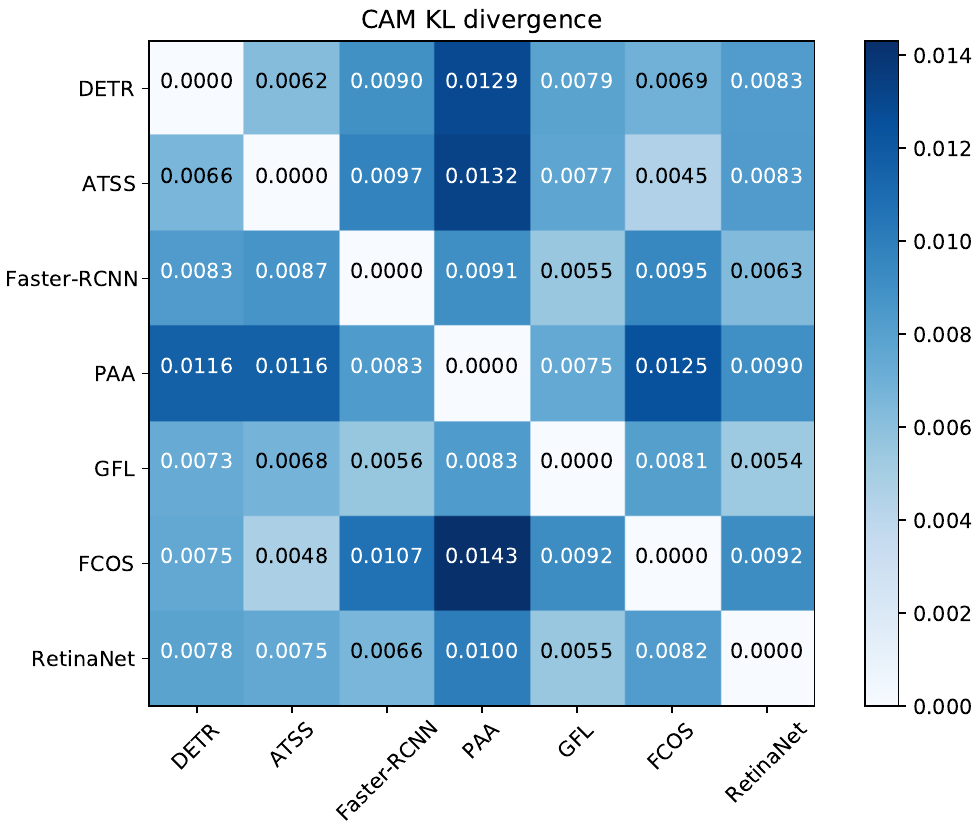}
    \vspace{-0.2cm}
    \caption{Distances among 7 various heads in our model with $K=6$.}
    \label{fig:mat_k7}
    \vspace{-4mm}
\end{figure}
\vspace{1mm}
\noindent\textbf{The number of customized positive queries.}
We compute the average ratio of positive samples in one-to-many label assignment to the ground-truth boxes.
For instance, the ratio is 18.7 for Faster-RCNN and 8.8 for ATSS on COCO dataset, indicating more than 8$\times$ extra positive queries are introduced when $K=1$.
% In fact, we simply set the maximum number of customized positive queries within a branch the same as the number of object queries.

\vspace{1mm}
\noindent\textbf{Effectiveness of collaborative one-to-many label assignments.}
% To verify the effectiveness of collaborative one-to-many label assignments, we compare our approach with Group-DETR ($3$ groups), which can be viewed as $K=2$ with two auxiliary one-to-one set matching branches.
To verify the effectiveness of our feature learning mechanism, we compare our approach with Group-DETR (3 groups) and $\mathcal{H}$-DETR.
First, we find $\mathcal{C}$o-DETR performs better than hybrid matching scheme~\cite{hybrid} while training faster and requiring less GPU memory in Table \textcolor{red}{6}.
As shown in Table \textcolor{red}{8}, our method ($K=1$) achieves 46.2\% AP, surpassing Group-DETR (44.6\% AP) by a large margin even without the customized positive queries generation.
More importantly, the IoF-IoB curve in Figure \textcolor{red}{2} demonstrates Group-DETR fails to enhance the feature representations in the encoder, while our method alleviates the poorly feature learning.

\vspace{1mm}
\noindent\textbf{Conflicts analysis.} 
We have defined the distance between head $H_i$ and head $H_j$, and the average distance of $H_i$ to measure the optimization conflicts in this study:
\begin{equation}
    \mathcal{S}_{i,j} =\frac{1}{|\mathbf{D}|} \sum_{\mathbf{I} \in \mathbf{D}} \mathrm{KL}(\mathcal{C}(H_i(\mathbf{I})), \mathcal{C}(H_j(\mathbf{I})),
\end{equation}
\begin{equation}
    \mathcal{S}_{i} =\frac{1}{2(K-1)} \sum_{j \neq i}^{K} (\mathcal{S}_{i,j}+\mathcal{S}_{j,i}),
\end{equation}
where $\mathrm{KL}$, $\mathbf{D}$, $\mathbf{I}$, $\mathcal{C}$ refer to KL divergence, dataset, the input image, and class activation maps (CAM)~\cite{cam}.
In our implementation, we choose the validation set COCO \texttt{val} as $\mathbf{D}$ and Grad-CAM as $\mathcal{C}$.
We use the output features of DETR encoder to compute the CAM maps.
More specifically, we show the detailed distances when $K=2$ and $K=6$ in Figure~\ref{fig:mat_k2} and Figure~\ref{fig:mat_k7}, respetively.
The larger distance metric of $\mathcal{S}_{i,j}$ indicates $H_i$ is less consistent to $H_j$ and contributes to the optimization inconsistency.

% \vspace{1mm}
% \noindent\textbf{Relations among different heads.}
% To better understand the diverse representations learned by multiple heads, we define the relation between head $H_i$ and head $H_j$ as:
% \begin{equation}
%     \mathcal{S}_{i,j} = \mathrm{KL}(\mathrm{CAM}(H_i), \mathrm{CAM}(H_j)),
% \end{equation}
% where $\mathrm{KL}$ denotes Kullback–Leibler divergence and $\mathrm{CAM}$ stands for the class activation maps obtained by Grad-CAM~\cite{cam}.
% We evaluate the relations for each image on COCO \texttt{val} and obtain the average values that are presented in Figure~\ref{fig:mat_k2} and Figure~\ref{fig:mat_k7}.
% First, we find the overall distance (average value of the relation matrix that excludes diagonal values) of $K=2$ ($0.0073$) is lower than the one of $K=6$ ($0.0084$).
% This indicates the CAM areas of different heads are more similar when $K=2$, which is consistent with our results.
% Then, the distance of our model with $K=6$ fluctuates significantly (from $0.0045$ to $0.0143$).
% Such head diversity leads to incompatible optimization targets and hurts the detection performance.

\section{More implementation details}
\vspace{1mm}
\noindent\textbf{One-stage auxiliary heads.}
Based on the conventional one-stage detectors, we experiment with various first-stage designs~\cite{atss,gfl,fcos,paa,retina} for the auxiliary heads.
First, we use the GIoU~\cite{giou} loss for the one-stage heads.
Then, the number of stacked convolutions is reduced from $4$ to $1$.
Such modification improves the training efficiency without any accuracy drop.
For anchor-free detectors, \eg, FCOS~\cite{fcos}, we assign the width of $8\times2^{j}$ and height of $8\times2^{j}$ for the positive coordinates with stride $2^{j}$.

\vspace{1mm}
\noindent\textbf{Two-stage auxiliary heads.}
We adopt the RPN and RCNN as our two-stage auxiliary heads based on the popular Faster-RCNN~\cite{faster} and Mask-RCNN~\cite{mask} detectors.
To make $\mathcal{C}$o-DETR compatible with various detection heads, we adopt the same multi-scale features (stride 8 to stride 128) as the one-stage paradigm for two-stage auxiliary heads.
Moreover, we adopt the GIoU loss for regression in the RCNN stage.

\vspace{1mm}
\noindent\textbf{System-level comparison on COCO.}
We first initialize the ViT-L backbone with EVA-02 weights.
Then we perform intermediate finetuning on the Objects365 dataset using $\mathcal{C}$o-DINO-Deformable-DETR for 26 epochs and reduce the learning
rate by a factor of 0.1 at epoch 24.
The initial learning rate is $2.5\times10^{-4}$ and the batch size is 224.
We choose the maximum size of input images as 1280 and randomly resize the shorter size to 480$-$1024.
Moreover, we use 1500 object queries and 1000 DN queries for this model.
Finally, we finetune $\mathcal{C}$o-DETR on COCO for 12 epochs with an initial learning rate of $5\times10^{-5}$ and drop the learning rate at the 8-th epoch by multiplying 0.1.
The shorter size of input images is enlarged to 480$-$1536 and the longer size is no more than 2400.
We employ EMA and train this model with a batch size of 64.

\vspace{1mm}
\noindent\textbf{System-level comparison on LVIS.}
In contrast to the COCO setting, we use $\mathcal{C}$o-DINO-Deformable-DETR++ to perform intermediate finetuning on the Objects365 dataset, as we find LSJ augmentation works better on the LVIS dataset.
A batch size of 192, an initial learning rate of $2\times10^{-4}$, and an input image size of 1280$\times$1280 are used.
We use 900 object queries and 1000 DN queries for this model.
During finetuning on LVIS, we arm it with an additional auxiliary mask branch and increase the input size to 1536$\times$1536.
Besides, we train the model without EMA for 16 epochs, where the batch size is set to 64, and the initial learning rate is set to $5\times10^{-5}$, which is reduced by a factor of 0.1 at the 9-th and 15-th epoch.

% \vspace{1mm}
% \noindent\textbf{Details of MixMIM-g.}
% Following the previous practice, we scale up the MixMIM~\cite{mixmim} to 1-billion parameters with configurations listed below: 
% \begin{itemize}[leftmargin=*]
%     \item channel numbers: $C=(384, 768, 1536, 3072)$, 
%     \item numbers of the attention heads: $H=(6, 12, 24, 48)$, 
%     \item numbers of blocks for each stage: $B = (2, 6, 24, 2)$.
% \end{itemize}

% \section{More results}
% As described in Table \ref{tab:coco_k1}, 
% we compare some $\mathcal{C}$o-DETR models ($K=1$) with other state-of-the-art frameworks.
% Our best model achieves the comparable performance of 58.3\% AP, and is able to outperform both DINO-Deformable-DETR (58.0\% AP) and $\mathcal{H}$-Deformable-DETR (57.9\% AP), further demonstrating the effectiveness of collaborative hybrid assignments training.

%% file: num_conv.tex
\begin{table}[t]
    \centering
    \footnotesize
    \resizebox{0.9\linewidth}{!}
    {
        \begin{tabular}{c|c|c|c|c|c|c}
        \shline
        \#convs & 0 & 1 & 2 & 3 & 4 & 5  \\
        \hline
        AP & 41.8 & \textbf{42.3} & 41.9 & 42.1 & \textbf{42.3} & 42.0 \\
        \shline
        \end{tabular}
    }
    \vspace{-2mm}
    \caption{\small{Influence of number of convolutions in auxiliary head.}
    }
    \label{tab:num_conv}
\end{table}

%% file: loss_head.tex
\begin{table}[t]
    \centering\setlength{\tabcolsep}{6pt}
    \footnotesize
    \renewcommand{\arraystretch}{1.2}
    \resizebox{0.9\linewidth}{!}
    {
        \begin{tabular}{l|c|c|c|c|c|c}
        \shline
        $\lambda_{1}$ & $\lambda_{2}$ & \#epochs  & AP & AP$_{S}$ & AP$_{M}$ & AP$_{L}$ \\
        \shline
        0.25 & 2.0 & 36 & 46.2 & 28.3 & 49.7 & 60.4 \\ 
        0.5 & 2.0 & 36 & 46.6 & 29.0 & 50.5 & 61.2 \\ 
        1.0 & 2.0 & 36 & \textbf{46.8} & \textbf{28.1} & \textbf{50.6} & \textbf{61.3} \\ 
        2.0 & 2.0 & 36 & 46.1 & 27.4 & 49.7 & 61.4 \\ 
        \hline
        1.0 & 1.0 & 36 & 46.1 & 27.9 & 49.7 & 60.9 \\ 
        1.0 & 2.0 & 36 & \textbf{46.8} & \textbf{28.1} & \textbf{50.6} & \textbf{61.3} \\ 
        1.0 & 3.0 & 36 & 46.5 & 29.3 & 50.4 & 61.4 \\ 
        1.0 & 4.0 & 36 & 46.3 & 29.0 & 50.1 & 61.0 \\ 
        \shline
        \end{tabular}
    }
    \vspace{-2mm}
    \caption{\small{Results of hyper-parameter tuning for $\lambda_{1}$ and $\lambda_{2}$.}
    \label{tab:loss_head}
    }
\end{table}
% \begin{table}[t]
%     \centering\setlength{\tabcolsep}{6pt}
%     \footnotesize
%     \renewcommand{\arraystretch}{1.2}
%     \caption{\small{Experimental results of $K$ varying from $1$ to $3$.}
%     \vspace{-3mm}
%     }
%     \label{tab:multi_head}
%     \resizebox{\linewidth}{!}
%     {
%         \begin{tabular}{c|l|c|c|c|c}
%         \shline
%         \multirow{2}{*}{$K$} & \multirow{2}{*}{Co-trained Head} & \multirow{2}{*}{Memory} & \multirow{2}{*}{Time} & \multicolumn{2}{c}{\#epochs}  \\
%         \cline{5-6}
%         & & & & 12 & 36 \\
%         \shline
%         $0$ & - & - & - & $37.1$ & $43.3$ \\ 
%         $1$ & ATSS & - & $42.3$ & $46.8$ \\ 
%         $2$ & ATSS + Faster-RCNN & - & $42.9$ & $46.5$ \\ 
%         $3$ & ATSS + Faster-RCNN + FCOS & - & $-$ & $-$ \\ 
%         % \hline
%         % Deformable-DETR & ATSS & - & - & $36$ & $46.8$ & $65.1$ & $51.5$ \\
%         %\rowcolor{blue!15}Deformable-DETR + ATSS & $36$ & $\underline{46.8}$ \\
%         \shline
%         \end{tabular}
%     }
% \end{table}